\documentclass[10pt, twocolumn]{article}
\usepackage[left=1.6cm,right=1.6cm,top=2cm,bottom=2cm]{geometry}
\usepackage{graphicx}
\usepackage{cite}
\usepackage[cmex10]{amsmath}

\usepackage{amssymb}
\usepackage{latexsym} 
\usepackage{tikz}
\usetikzlibrary{decorations.markings}

\usepackage{booktabs}

\begin{document}
\title{Image Segmentation and Restoration Using Parametric Contours With Free Endpoints}
\author{Heike~Benninghoff
\thanks{Deutsches Zentrum f\"ur Luft- und Raumfahrt (DLR), 82234 We\ss ling, Germany, email: heike.benninghoff@dlr.de.} 
~and~Harald~Garcke
\thanks{Fakult\"at f\"ur Mathematik, Universit\"at Regensburg, 93040 Regensburg, Germany, email: harald.garcke@ur.de.}}

\date{}
\maketitle

\begin{abstract}
In this paper, we introduce a novel approach for active contours with free endpoints. A scheme is presented for image segmentation and restoration based on a discrete version of the Mumford-Shah functional where the contours can be both closed and open curves. Additional to a flow of the curves in normal direction, evolution laws for the tangential flow of the endpoints are derived. Using a parametric approach to describe the evolving contours together with an edge-preserving denoising, we obtain a fast method for image segmentation and restoration. The analytical and numerical schemes are presented followed by numerical experiments with artificial test images and with a real medical image. 
\end{abstract}


\section{Introduction}
\label{sec:intro}

This article addresses important classical problems in image processing: image segmentation, edge detection and image restoration. 

Image segmentation aims at partitioning a given image into its constituent parts, also called regions or phases. A segmentation of an image can be given by a set of region boundaries and edges. Different types of edges can occur in images: edges can be boundaries of objects and separate these objects from their background or from each other. But edges can also end inside the image at a location where no other edge continues. 

Boundaries of objects can be modeled with so-called interface curves. Non-interface curves are curves which do not separate two different regions in the image. Such curves have one or two so-called free endpoints. 

Image restoration aims at reducing or removing noise which affects a given image. Typically, a blurring of the sharp edges in the image should be prevented when smoothing an image. This results in the need of an edge preserving image denoising method. 

Image segmentation including edge detection can be performed with active contours (also called snakes), first proposed by Kass, Witkin, and Terzopoulos \cite{Kass88} in 1988. Since this time, the popular method is applied and further developed by many authors, e.g. \cite{Cohen91, Ronfard94, Caselles97, Malladi95, Kichenassamy96, Araki97, Paragios00, Chan00, Chan01}. Using active contours, a curve evolves in order to minimize a given energy functional. The energy functional should be designed such that a minimizing curve matches with the region boundaries or edges in the image. 

The Mumford-Shah functional \cite{Mumford89} can be used for both image segmentation and image restoration. A pair $(\Gamma,u)$ should be found which minimizes the Mumford-Shah energy, where $\Gamma$ is a set of curves and $u$ is a piecewise smooth function with possible discontinuities across $\Gamma$. Having found a solution $(\Gamma,u)$, a segmentation of the image is given by the set of object boundaries and edges $\Gamma$, and a denoised version of the image is given by the piecewise smooth approximation $u$. 

An important variant of the Mumford-Shah problem is the restriction to piecewise \emph{constant} image approximations $u$, the so-called minimal partition problem \cite{Chan01}. However if edges with free endpoints, also called crack-tips \cite{Mumford89}, occur, the piecewise constant approximation will not be applicable. 

It is also possible to approximate the Mumford-Shah functional by a sequence of simpler elliptic variational problems as introduced by Ambrosio and Tortorelli \cite{AmbrosioTortorelli90}. They replaced the curve $\Gamma$ by a 2D function for which a phase field type energy is added to the functional. 

Image segmentation and restoration are classical areas in image analysis, see \cite{Kass88, Mumford89, Rudin1992, Ronfard94}, but still significant in more present research, see e.g. \cite{Chan01,Chan01_2, Tsai01, Yu02, Benes04, Dogan08, Chung2009, Mille2009, Arbelaez11, Chambolle2012} to mention some selected works. There is also a variety of related image processing tasks like object detection \cite{Paragios00, Fergus03} or pattern recognition \cite{Bezdek2005}), feature extraction \cite{Nixon02} and anomaly detection \cite{Chandola2009}.  

The image segmentation method, considered and developed in this article, also uses the evolution of curves. The resulting evolution equations, derived from the Mumford-Shah functional, can be written as parabolic partial differential equations for a parametrization of the curves $\Gamma$. The restoration is performed by solving a diffusion equation for $u$, also derived from the Mumford-Shah model. By using the location of the curves $\Gamma$, we obtain an edge-preserving smoothing. 

Open active contours, i.e. active contours with free endpoints, are also considered by \cite{Kimmel2003}, where the authors propose a method for detection of open boundaries based on an edge detector which uses the image gradient. Here, we consider approaches based on the Mumford-Shah model. Using convex relaxation approaches, global minimizers of the Mumford-Shah functional are determined in \cite{Pock2009}. The method can also handle free endpoints. In \cite{Schaeffer13}, the level set method is used for evolving curves with free endpoints. However, two level set functions and artificial regions are needed to describe a curve with free endpoints. 

During the evolution of curves, topology changes like splitting or merging can occur, since the number and the topology type of edges and region boundaries is often not known in advance. Using indirect methods like level set and phase field techniques, topology changes are handled automatically. It is often argued that the inability to change the topology of curves is the main disadvantage of parametric methods like the original snake model \cite{Kass88}. In this paper, we extend an efficient method to detect and perform topology changes (presented in \cite{Benninghoff2014a} and based on the original idea of \cite{Balazovjech12, MikulaUrban12}), such that also topology changes of curves with free endpoints can be handled. 

The objective of this article is to solve the Mumford-Shah problem including curves with free endpoints with a parametric approach. The method we propose is based on a parametrization of the evolving curves. We show how a method developed for interface curves \cite{Benninghoff2014a} can be extended for curves with free endpoints. With the presented concept for image segmentation and restoration, we can easily process images with both open and closed edges. Our method is very efficient from a computational point of view, since the curve evolution problem is a one-dimensional problem and no artificial regions have to be used compared to \cite{Schaeffer13}.


\section{Image Processing with Parametric Contours}
\label{sec:im_proc_methods}
Let $u_0: \Omega\rightarrow \mathbb{R}$ be an image function describing for each point in the image domain $\Omega \subset \mathbb{R}^2$ the intensity of the image. 

The Mumford-Shah method \cite{Mumford89} for optimal approximation of images aims at finding a set of curves $\Gamma = \Gamma_1 \cup \ldots \Gamma_{N_C}$ and a piecewise smooth function $u:\Omega \rightarrow \mathbb{R}$ with possible discontinuities across $\Gamma$ approximating the original image $u_0$. The energy to be minimized is 
\begin{equation}
E(u, \Gamma) = \sigma |\Gamma| + \int_{\Omega\setminus \Gamma} \|\nabla u\|^2 \,\mathrm{d}x + \lambda \int_\Omega (u_0-u)^2 \,\mathrm{d}x,
\label{eq:mumford_shah}
\end{equation}
where $\sigma, \lambda > 0$ are weighting parameters and $|\Gamma|$ denotes the total length of the curves in $\Gamma$. 

A minimizer of the Mumford-Shah functional provides (i) a restoration of the possible noisy original image by a piecewise smooth approximation $u$ and (ii) a segmentation of the image given by a union of curves $\Gamma$ representing the set of edges in the image. The curves belonging to $\Gamma$ can be sharp edges where the image function rapidly changes, but they can also be so-called weak edges where the image function smoothly changes its value, see \cite{Chan01}.   

\begin{figure}[t]
	\centering
		\includegraphics[viewport = 150 270 435 560, width=0.2\textwidth]{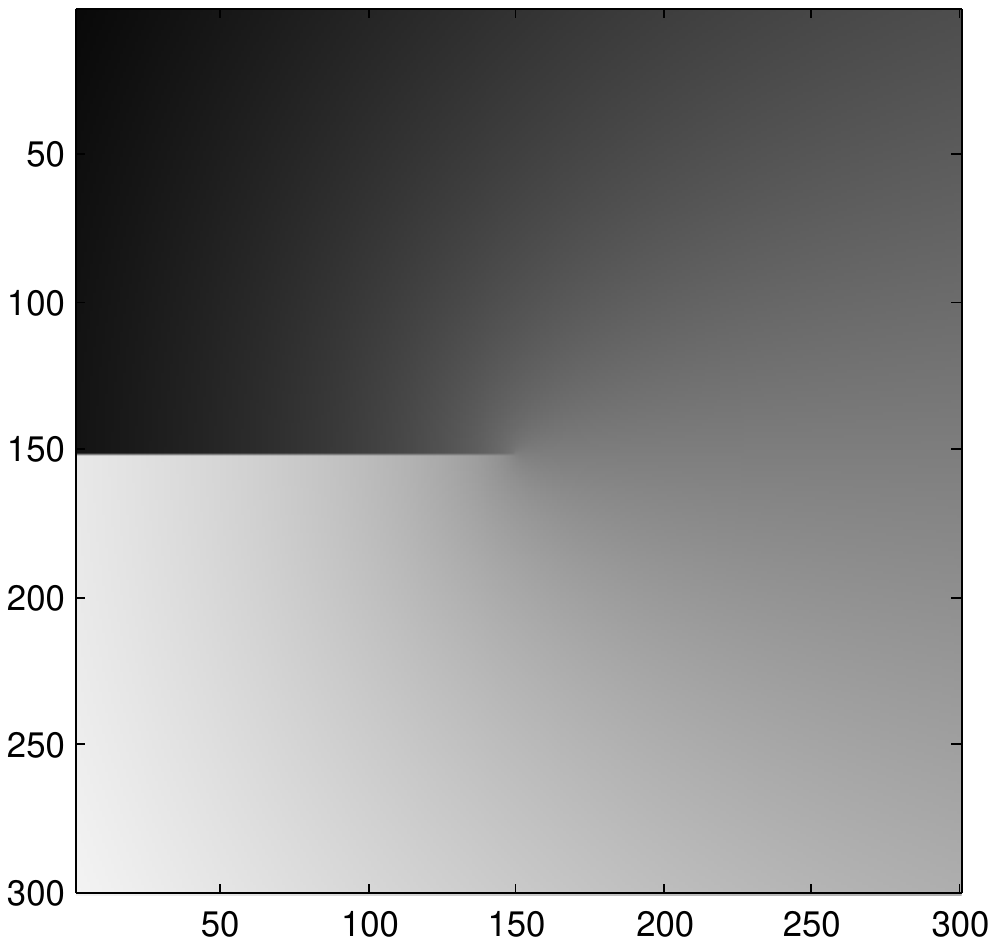}
	\caption{Image containing an edge with a free endpoint.}
	\label{fig:example_image_free_endpoints}
\end{figure}

The contours $\Gamma_i$, $i=1, \ldots, N_C$, may be closed contours with $\partial \Gamma = \emptyset$, or open contours with two endpoints. The endpoints may lie on the image boundary $\partial \Omega$, may belong to triple junctions where three curves meet, or may be free endpoints,  cf. the conjecture of Mumford and Shah \cite{Mumford89}. In the latter case, the endpoint is a point inside the image domain, where no other curve continues. Figure~\ref{fig:example_image_free_endpoints} shows an image where an edge occurs which terminates near the image center. The edge can be represented by a curve with one endpoint located at the left image boundary $\partial \Omega$ and one endpoint being a free endpoint, located close to the image center. 

In \cite{Benninghoff2014a}, we proposed a parametric method for image segmentation with piecewise constant image approximations $u$ and interface curves $\Gamma_1, \ldots, \Gamma_{N_C}$, each separating two regions. There, we considered a decomposition of the image in $N_R$ regions $\Omega_1, \ldots, \Omega_{N_R}$ separated by curves $\Gamma_i$, $i=1, \ldots, N_C$, and approximations $u_{|\Omega_k}=c_k\in\mathbb{R}$. In that case, the functional \eqref{eq:mumford_shah} reduces to
\begin{equation}
E(\Gamma, c_1, \ldots, c_{N_R}) = \sigma |\Gamma| + \lambda \sum_{k=1}^{N_R}\int_{\Omega_k} (u_0-c_k)^2 \,\mathrm{d}x,
\label{eq:mumford_shah_pw_constant}
\end{equation}
see \cite{Chan01}. Using methods from the theory of calculus of variations, the following evolution equation can be derived for time-dependent curves $\Gamma_i(t)$, $t \in [0,T]$: 
\begin{equation}
(V_n)_i = \sigma \kappa_i + F_i, \quad\,i=1, \ldots, N_C,
\label{eq:evolution_eq_pw_constant}
\end{equation}
where $(V_n)_i$ is the normal velocity of $\Gamma_i(t)$, $\kappa_i$ is the curvature, and $F_i$ is given by
\begin{equation}
F_i(\,.\,,t) = \lambda [ (u_0 - c_{k^+(i)}(t))^2 -  (u_0 - c_{k^-(i)}(t))^2 ],  
\label{eq:F_pw_constant}
\end{equation}
with $t \in [0,T]$. The indices $k^+(i), k^-(i) \in \{1, \ldots, N_R\}$ denote the two regions which are separated by $\Gamma_i(t)$. The coefficients $c_k(t)$, $k=1, \ldots, N_R$, are the mean of $u_0$ in the region $\Omega_k(t)$. 

In practice, the segmentation problem can be solved in a two-step approach. For discrete time steps $t\geq 0$, the coefficients $c_k$ are computed using the current set of curves. This is followed by an update of the curves $\Gamma(t) \rightarrow \Gamma(t + \Delta t)$, performed by solving the evolution equation \eqref{eq:evolution_eq_pw_constant}.

For some images, the piecewise constant approximation is not applicable, see the exemplary image in Figure~\ref{fig:example_image_free_endpoints}. For such images, the image domain cannot be decomposed in regions separated by interface curves. 

Consequently, we modify the two-step approach of \cite{Benninghoff2014a}, such that also non-interface curves with free endpoints can be dealt with. In the first step, we will solve a diffusion equation in the image domain resulting in a piecewise smooth approximation $u$. Instead of using the coefficients $c_k$, we will consider for $\vec p \in \Gamma_i(t)$ the limit $u^\pm(\vec p) = \lim_{\epsilon \rightarrow 0, \epsilon > 0} u(\vec p \pm \epsilon \vec \nu_i(\vec p))$, where $\vec \nu_i$ is a normal vector field on $\Gamma_i(t)$.  Having computed $u$, we solve the evolution equation \eqref{eq:evolution_eq_pw_constant} with a modified external term $F_i$, using  $u^\pm$ instead of constants $c_{k^\pm(i)}$.   

Before, presenting further details, we first consider the regularity of a solution of the Mumford-Shah problem at the free endpoint. Fixing the curves $\Gamma$, let $u$ denote the minimizer of the Mumford-Shah energy \eqref{eq:mumford_shah}. At free endpoints, problems concerning the regularity of $u$ occur, cf. \cite{Mumford89}. Expressed in polar coordinates $(r,\phi)$ centered at the free endpoint, the solution $u$ is of the form
\begin{equation}
u(r, \phi) = c\, r^{1/2} \sin(\frac12 (\phi - \phi_0) ) + \hat{v}(r, \phi),
\end{equation}
where $\hat{v}$ is a $C^1$-function and $c, \phi_0$ are constants, see \cite{Aubert06}. 

For image segmentation, we later need to solve the problem on a discrete set: Let $\Omega^h$ be a rectangular grid of nodes covering $\Omega$ with grid size $h>0$.  We replace the second integral on the right hand side of \eqref{eq:mumford_shah} by a sum containing difference quotients of the form
\begin{equation}
\nabla_h^i u(\vec z) = \frac1h (u(\vec z + h \vec e_i) - u(\vec z)), \quad \vec z \in \Omega^h, 
\end{equation}
where $\vec e_i \in \mathbb{R}^2$ are the standard basis vectors of $\mathbb{R}^2$, $i=1,2$. For image segmentation applications, we choose the pixel grid, i.e. we use $h=1$. For the approximating sum, we have to exclude terms where the line $[\vec z, \vec z + h \vec e_i]$ intersects with the curve $\Gamma$. 

Instead of the original Mumford-Shah functional \eqref{eq:mumford_shah}, we thus consider the energy
\begin{align}
&E^h(\Gamma,u) = \sigma |\Gamma | +  \sum_{\substack{\vec z \in \Omega^h \\\text{s.t. }\vec z + h \vec e_2 \in \Omega^h}} (1-\alpha_x(\vec z)) (\nabla_h^2 u(\vec z) )^2 +  \nonumber \\
&  + \sum_{\substack{\vec z \in \Omega^h \\\text{s.t. }\vec z + h \vec e_1 \in \Omega^h}}(1-\alpha_y(\vec z)) (\nabla_h^1 u(\vec z))^2  + \lambda \int_\Omega (u_0-u)^2 \,\mathrm{d}x,
\label{eq:mumford_shah_discrete}
\end{align}
where $\alpha_x(\vec z), \alpha_y(\vec z) \in [0,1]$ are scalar terms. If $[\vec z, \vec z + h \vec e_1]$ intersects with $\Gamma$, $\alpha_y(\vec z)$ is set to $1$.

\subsection{Example}
We consider one single open curve $\Gamma$. Let $\vec x: [0,1] \rightarrow \mathbb{R}^2$ with $\vec x([0,1])=\Gamma$ be a parameterization of the curve. Let $\vec x(0)$ be a free endpoint and let $\vec x(1)$ intersect with the image boundary. 

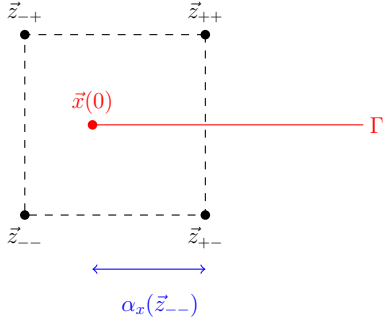
\begin{figure}[t]
\begin{center}
\begin{tikzpicture}[scale=0.6,transform shape]

\draw[dashed] (2,2) -- (2,-2) -- (-2,-2) -- (-2,2) -- (2,2); 

\fill ( 2, 2) circle (3pt);
\fill ( 2,-2) circle (3pt);
\fill (-2,-2) circle (3pt);
\fill (-2, 2) circle (3pt);
\fill[red] (-0.5, 0) circle (3pt);
\draw[red] (-0.5,0) -- (5.5,0); 
\draw[red] (5.8,0) node{\Large $\Gamma$};

\draw( 2, 2.5) node{\Large $\vec z_{++}$};
\draw( 2,-2.5) node{\Large $\vec z_{+-}$};
\draw(-2,-2.5) node{\Large $\vec z_{--}$};
\draw(-2, 2.5) node{\Large $\vec z_{-+}$};
\draw[red] (-0.5, 0.5) node{\Large $\vec x(0)$};

\draw[<->,blue] (-0.5, -3.2) --  (2,-3.2); 
\draw[blue](1,-4.0) node{\Large $\alpha_x(\vec z_{--})$};

\end{tikzpicture}
\end{center}
\caption{Illustration of the pixel grid close to the free endpoint.}
\label{fig:grid_at_freeendpoint}
\end{figure}

Figure~\ref{fig:grid_at_freeendpoint} visualizes a possible situation near the free endpoint $\vec x(0)$. Let $\vec z_{++}$, $\vec z_{+-}$, $\vec z_{--}$, $\vec z_{-+}$ denote the four grid points around $\vec x(0)$ as shown in Figure~\ref{fig:grid_at_freeendpoint}. In this example, the tangential vector of the curve at $\vec x(0)$ is $\vec\tau(0) = \vec x_s(0) = \vec e_1$, where $s$ denotes the arc-length of the curve. 

Considering $\vec z = \vec z_{+-}$, the line $[\vec z_{+-}, \vec z_{++}]$ and $\Gamma$ intersect. Thus, $\alpha_x(\vec z_{+-})$ is set to $1$. For $\vec z = \vec z_{--}$, we define a factor 
\begin{equation}
\alpha_x(\vec z_{--}) := \frac1h \left( (\vec z_{+-})_1 - (\vec x(0))_1 \right),
\end{equation}
where $(\,.\,)_i$ denotes the $i$-th component of a vector, $i=1,2$. The factor $\alpha_x(\vec z_{--})$ describes how far the curve has entered the square given by $\vec z_{++}$, $\vec z_{+-}$, $\vec z_{--}$, $\vec z_{-+}$. 

For $\vec z \in \Omega^h$, $\vec z \neq \vec z_{--}$, we set $\alpha_x(\vec z)=0$ if $[\vec z, \vec z + h \vec e_2] \cap \Gamma =\emptyset$ and $\alpha_x(\vec z)=1$ else. The factor  $\alpha_y(\vec z)$ is defined similarly. 

We now want to vary $\Gamma$ in direction $-\vec \tau(0)$ at $\vec x(0)$. We consider a second curve $\Gamma^\epsilon$, $\epsilon>0$, with a parameterization $\vec x^\epsilon$, such that $\vec x^\epsilon(0) = \vec x(0) - \epsilon \vec\tau(0)$. We can assume that $\epsilon$ is small enough, such that $\vec x^\epsilon(0)$ is still inside the square given by $\vec z_{++}$, $\vec z_{+-}$, $\vec z_{--}$, $\vec z_{-+}$. 

The energy difference is 
\begin{align*}
E^h(\Gamma^\epsilon,u) - E^h(\Gamma,u) =& \sigma \epsilon  +  (1 - \alpha_x(\vec z_{--})- \epsilon) (\nabla_h^2 u(\vec z_{--}))^2 \\
& - (1 - \alpha_x(\vec z_{--})) (\nabla_h^2 u(\vec z_{--}))^2  \\
=& \sigma \epsilon  - \epsilon   (\nabla_h^2 u(\vec z_{--}))^2.
\end{align*}
The energy will decrease if 
\begin{equation}
\sigma <  (\nabla_h^2 u(\vec z_{--}))^2.
\label{eq:condition_decrease}
\end{equation}
Thus a motion of a curve in direction $-\vec\tau(0)=-\vec e_1$ at the free endpoint $\vec x(0)$ requires that the square of the difference quotient of $u$ in $\vec e_2$-direction at $\vec z_{--}$ is sufficient large compared to the weighting parameter $\sigma$ of the length term in the energy \eqref{eq:mumford_shah_discrete}.  

\subsection{General Case}
We consider a curve with one or two free endpoints with tangential vector $\vec\tau(\rho)$ at the free endpoint $\vec x(\rho)$,  $\rho \in \{0,1\}$. We define the factors $\alpha_x$ and $\alpha_y$ as follows: Let $\vec z_{++}(\rho), \vec z_{+-}(\rho), \vec z_{--}(\rho), \vec z_{-+}(\rho)$ denote the four grid points around $\vec x(\rho)$.

If $\rho = 0$ and $\vec \tau(0) \,.\,\vec e_1 \geq 0$ or if $\rho = 1$ and $\vec \tau(1) \,.\,\vec e_1 < 0$, we define $\vec z^{\rho,1} := \vec z_{--}(\rho)$ and set 
\begin{equation*}
\alpha_x(\vec z^{\rho,1}) := 1 - \frac1h \left((\vec x(\rho))_1 - (\vec z^{\rho,1})_1 \right).
\end{equation*}

If $\rho = 0$ and $\vec \tau(0) \,.\,\vec e_1 < 0$ or if $\rho = 1$ and $\vec \tau(1) \,.\,\vec e_1 \geq 0$, we define $\vec z^{\rho,1} := \vec z_{+-}(\rho)$ and set
\begin{equation*}
\alpha_x(\vec z^{\rho,1}) := 1 - \frac1h \left((\vec z^{\rho,1})_1 - (\vec x(\rho))_1 \right).
\end{equation*}

If $\rho = 0$ and $\vec \tau(0) \,.\,\vec e_2 \geq 0$ or if $\rho = 1$ and $\vec \tau(1) \,.\,\vec e_2 < 0$, we define $\vec z^{\rho,2} := \vec z_{--}(\rho)$ and set
\begin{equation*}
\alpha_y(\vec z^{\rho,2}) := 1 - \frac1h \left((\vec x(\rho))_2 - (\vec z^{\rho,2})_2 \right).
\end{equation*}

If $\rho = 0$ and $\vec \tau(0) \,.\,\vec e_2 < 0$ or if $\rho = 1$ and $\vec \tau(1) \,.\,\vec e_2 \geq 0$, we define $\vec z^{\rho,2} := \vec z_{-+}(\rho)$ and set 
\begin{equation*}
\alpha_y(\vec z^{\rho,2}) := 1 - \frac1h \left((\vec z^{\rho,2})_2 - (\vec x(\rho))_2 \right).
\end{equation*}

Using these definitions, we define the following factors for $\vec z \in \Omega^h$:
\begin{equation*}
\alpha_x(\vec z) = \left\{
\begin{array}{ll}
1, & \text{if } \, [\vec z, \vec z + h\vec e_2]\cap\Gamma \neq \emptyset, \, \text{ and} \\
   &  \vec z \neq \vec z^{\rho,1}, \,\rho\in\{0,1\}, \\
\alpha_x(\vec z^{\rho,1}), & \text{if }\, \vec z = \vec z^{\rho,1}, \rho\in\{0,1\}, \\
0, & \text{else}. 
\end{array}
\right.
\end{equation*}
and
\begin{equation*}
\alpha_y(\vec z) = \left\{
\begin{array}{ll}
1, & \text{if } \, [\vec z, \vec z + h\vec e_1]\cap\Gamma \neq \emptyset, \, \text{ and} \\
   & \vec z \neq \vec z^{\rho,2}, \, \,\rho\in\{0,1\}, \\
\alpha_y(\vec z^{\rho,2}), & \text{if }\, \vec z = \vec z^{\rho,2}, \rho\in\{0,1\}, \\
0, & \text{else}. 
\end{array}
\right.
\end{equation*}

For minimizing \eqref{eq:mumford_shah_discrete}, we propose the following approach: 

Assume the case of one curve $\Gamma$ with two free endpoints parameterized by $\vec x: [0,1] \rightarrow \mathbb{R}^2$.
In the first step, we fix $u$ in \eqref{eq:mumford_shah_discrete} and consider for $\vec \eta:[0,1] \rightarrow \mathbb{R}^2$ a variation of $\vec x$ of the form $\vec x + \epsilon \vec \eta$, $\epsilon >0$.  

Let $\Gamma^{\epsilon, \eta}$ denote the image of $\vec x + \epsilon \vec \eta$. We use the notation $E^h(\vec\eta) := E^h(\Gamma^{\epsilon,\eta},u)$ and compute
\begin{align*}
\left.\frac{\mathrm{d}}{\mathrm{d}\epsilon}\right|_{\epsilon=0} E^h(\vec \eta) =& \left.\frac{\mathrm{d}}{\mathrm{d}\epsilon}\right|_{\epsilon=0}E^h(\Gamma^{\epsilon,\eta},u) \\
=& - \sigma \int_\Gamma \vec x_{ss} \,.\,\vec \eta\,\mathrm{d}s - \int_\Gamma F \,\vec\nu\,.\,\vec\eta\,\mathrm{d}s + \\
 & + \sigma \vec x_s(1)\,.\,\vec\eta(1) - \sigma \vec x_s(0)\,.\,\vec\eta(0) \\
 & - \mathrm{sign}(\vec \tau(1) \,.\, \vec e_1) \vec \eta(1)\,.\,\vec e_1 (\nabla_h^2 u(\vec z^{1,1}))^2   \\
 & - \mathrm{sign}(\vec \tau(1) \,.\, \vec e_2) \vec \eta(1)\,.\,\vec e_2 (\nabla_h^1 u(\vec z^{1,2}))^2   \\
 & + \mathrm{sign}(\vec \tau(0) \,.\, \vec e_1) \vec \eta(0)\,.\,\vec e_1 (\nabla_h^2 u(\vec z^{0,1}))^2   \\
 & + \mathrm{sign}(\vec \tau(0) \,.\, \vec e_2) \vec \eta(0)\,.\,\vec e_2 (\nabla_h^1 u(\vec z^{0,2}))^2. 
\end{align*}
For this computation, integration by parts and a transport theorem are applied. Further, $\vec\nu$ is a normal vector field on $\Gamma$ such that the pair $(\vec x_s, \vec \nu)$ is a positive oriented basis of $\mathbb{R}^2$ and $F$ is defined as the jump 
\begin{equation}
F=\lambda[(u_0 - u^+)^2 - (u_0 - u^-)^2].
\end{equation}

We define the following inner product for functions $\vec\eta,\, \vec\chi: [0,1]\rightarrow \mathbb{R}^2$:
\begin{equation}
\left(\vec\eta,\vec\chi\right)_{2,\Gamma,\partial\Gamma} :=  \int_\Gamma \vec\eta \,.\, \vec\chi \,\mathrm{d}s + \vec\eta(1)\,.\, \vec \chi(1) + \vec\eta(0)\,.\, \vec \chi(0).
\end{equation}

Now, we consider a family of curves $\Gamma(t)$, $t\in[0,T]$. Let $\vec x: [0,1] \times [0,T] \rightarrow \mathbb{R}^2$ be a mapping such that $\vec x(\,.\,,t)$ is a parameterization of $\Gamma(t)$, $t\in[0,T]$. We call $\vec x$ a solution of the gradient flow equation, if 
\begin{equation}
\left(\vec x_t,\vec\eta\right)_{2,\Gamma,\partial\Gamma} = - \left.\frac{\mathrm{d}}{\mathrm{d}\epsilon}\right|_{\epsilon=0} E^h(\vec \eta)
\label{eq:gradient_flow}
\end{equation}
holds for all $\vec \eta:[0,1]\rightarrow \mathbb{R}^2$. 

In the following, we consider particular choices of functions $\vec \eta$ and derive evolution equations for the curve. First, we consider $\vec \eta = \eta_0 \, \vec \nu$ for a scalar function $\eta_0:[0,1]\rightarrow \mathbb{R}$ with $\eta_0(0)=\eta_0(1)=0$. This provides
\begin{equation*}
\int_{\Gamma(t)} \vec x_t \,.\,\vec\nu \,\eta_0 \,\mathrm{d}s = \int_{\Gamma(t)}  (\sigma \vec x_{ss} \,.\, \vec \nu + F)\, \eta_0\,\mathrm{d}s.
\end{equation*}
Since $\eta_0$ is arbitrary chosen (with value $0$ at the endpoints), we conclude the following equation for the normal velocity of the curve:
\begin{subequations}
\label{eq:parametric_scheme_free_endpoints}
\begin{equation}
V_n := \vec x_t \,.\,\vec\nu = \sigma \kappa + F,
\label{eq:twodim_free_endpoints_eq_Vn}
\end{equation}
using the identity 
\begin{equation}
\kappa \vec \nu = \vec x_{ss}.
\label{eq:curvature_identity}
\end{equation}

Next, we choose $\vec \eta = \eta_0 \vec \tau$, where $\eta_0:[0,1]\rightarrow \mathbb{R}$ is a scalar function with $\eta_0(0)\neq 0$ and $\eta_0(1)=0$, i.e. $\vec \eta(0)=\vec\tau(0)\eta_0(0)$ and $\vec\eta(1)=\vec 0$. Inserting $\vec \eta$ in \eqref{eq:gradient_flow}, and using \eqref{eq:twodim_free_endpoints_eq_Vn}, \eqref{eq:curvature_identity} and $\vec\tau = \vec x_s$, leads to
\begin{align*}
&\vec x_t(0) \,.\,\vec \tau(0) \eta_0(0) =  \sigma \eta_0(0) \\
& \,\,- \mathrm{sign}(\vec \tau(0) \,.\, \vec e_1) \vec\tau(0)\eta_0(0)\,.\,\vec e_1 (\nabla_h^2 u(\vec z^{0,1}))^2 + \\
& \,\,- \mathrm{sign}(\vec \tau(0) \,.\, \vec e_2) \vec\tau(0)\eta_0(0)\,.\,\vec e_2 (\nabla_h^1 u(\vec z^{0,2}))^2.
\end{align*}
Since $\eta_0(0)$ is arbitrary and $\mathrm{sign}(\vec \tau(0) \,.\, \vec e_i) \vec\tau(0)\,.\,\vec e_i = |\vec\tau(0)\,.\,\vec e_i|$ for $i=1,2$, we conclude for the tangential velocity 
\begin{align}
V_\mathrm{tan}(0) :=& \vec x_t(0) \,.\,\vec \tau(0) \nonumber\\
 =&  \sigma - |\vec\tau(0)\,.\,\vec e_1| (\nabla_h^2 u(\vec z^{0,1}))^2  \nonumber\\
  &         - |\vec\tau(0)\,.\,\vec e_2| (\nabla_h^1 u(\vec z^{0,2}))^2.
\label{eq:Vtan0}
\end{align}

Choosing $\eta_0(0)=0$ and $\eta_0(1)\neq 0$, we can derive the following equation for the tangential velocity in $\vec x(1)$:
\begin{align}
V_\mathrm{tan}(1) :=& \vec x_t(1) \,.\,\vec \tau(1) \nonumber\\ 
 =& -\sigma + |\vec\tau(1)\,.\,\vec e_1| (\nabla_h^2 u(\vec z^{1,1}))^2  \nonumber\\
  &         + |\vec\tau(1)\,.\,\vec e_2| (\nabla_h^1 u(\vec z^{1,2}))^2.
\label{eq:Vtan1}
\end{align}

Similarly, choosing $\vec \eta = \eta_0 \vec \nu$, provides the following equations for the normal velocity at the free endpoints:
\begin{align}
V_\mathrm{n}(0) :=& \vec x_t(0) \,.\,\vec \nu(0) \nonumber\\ 
 =&         - \mathrm{sign}(\vec \tau(0) \,.\, \vec e_1) \vec \nu(0)\,.\,\vec e_1 (\nabla_h^2 u(\vec z^{0,1}))^2  \nonumber\\
  &         - \mathrm{sign}(\vec \tau(0) \,.\, \vec e_2) \vec \nu(0)\,.\,\vec e_2 (\nabla_h^1 u(\vec z^{0,2}))^2,
\label{eq:Vn0}
\end{align}
\begin{align}
V_\mathrm{n}(1) :=& \vec x_t(0) \,.\,\vec \nu(0) \nonumber\\ 
 =&         + \mathrm{sign}(\vec \tau(1) \,.\, \vec e_1) \vec \nu(1)\,.\,\vec e_1 (\nabla_h^2 u(\vec z^{1,1}))^2  \nonumber\\
  &         + \mathrm{sign}(\vec \tau(1) \,.\, \vec e_2) \vec \nu(1)\,.\,\vec e_2 (\nabla_h^1 u(\vec z^{1,2}))^2.
\label{eq:Vn1}
\end{align}
\end{subequations}

The curve $\Gamma$ will grow locally at $\vec x(0)$, if the curve moves in direction $-\vec\tau(0)$. In this case $V_\mathrm{tan}(0) = \vec x_t(0) \,.\,\vec\tau(0) < 0$. Therefore,
\begin{equation}
 \sigma < |\vec\tau(0)\,.\,\vec e_1| (\nabla_h^2 u(\vec z^{0,1}))^2 + |\vec\tau(0)\,.\,\vec e_2| (\nabla_h^1 u(\vec z^{0,2}))|^2
\label{eq:free_endpoints_inequality1}
\end{equation}
has to be satisfied such that the curve length increases. For the exemplary case $\vec\tau(0) = \vec e_1$, the condition reduces to  \eqref{eq:condition_decrease}, i.e. to the condition from the introductory example.

The curve $\Gamma(t)$ will grow at $\vec x(1)$, if the curve moves in direction $\vec\tau(1)$ leading to $V_\mathrm{tan}(1) > 0$. Therefore, the inequality 
\begin{equation}
\sigma < |\vec\tau(1)\,.\,\vec e_1| (\nabla_h^2 u(\vec z^{1,1}))^2 + |\vec\tau(1)\,.\,\vec e_2| (\nabla_h^1 u(\vec z^{1,2}))^2
\label{eq:free_endpoints_inequality2}
\end{equation}
has to be satisfied. 

Since the term $\sigma |\Gamma|$ in the energy \eqref{eq:mumford_shah_discrete} penalizes the length of the curve, a curve can only grow in direction $-\vec\tau(0)$ or $\vec\tau(1)$, if the derivative terms $(\nabla_h^{i}u)^2$, $i=1,2$, are large compared to $\sigma$. 

The scheme \eqref{eq:parametric_scheme_free_endpoints} describes the motion of the curve. For $N_C$ curves $\Gamma_1, \ldots, \Gamma_{N_C}$, we can solve \eqref{eq:parametric_scheme_free_endpoints} for each curve. For a closed curve $\Gamma_i$, only the normal velocity \eqref{eq:twodim_free_endpoints_eq_Vn} with the relation \eqref{eq:curvature_identity} needs to be considered, on noting that $\vec x_i(0)=\vec x_i(1)$. In the case of triple junctions and intersections with the image boundary, additional conditions for the involved endpoints have to be considered. If triple junctions occur, the evolution equations for the corresponding three curves which meet at the junction are coupled. The cases with triple junctions and boundary intersection points are described in \cite{Benninghoff2014a} in detail. 

We alternately solve the scheme of evolution equations \eqref{eq:parametric_scheme_free_endpoints} and recompute the approximating function $u$ using the updated curve set. The function $u$ is obtained by solving a diffusion equation on $\Omega^h$. We note that \eqref{eq:mumford_shah_discrete} is formulated for a discrete set $\Omega^h$. We will describe in the next section, how $u$ is computed numerically.

%

%

\section{Numerical Approximation}
\label{sec:numerics}
\subsection{Numerical Solution of the Evolution Equations}
For computing the position of the evolving curves $\Gamma$ numerically, we consider a decomposition of the interval $[0,1]$ of the form 
$0=q_0^i < q_1^i < \ldots < q_{N_i}^i = 1$, for $i=1, \ldots, N_C$. If $\Gamma_i$ is a closed curve, we make use of the periodicity $N_i=0$, $N_i+1=1$, $-1=N_i-1$, etc. 

Further, let $0=t_0 < t_1 < \ldots < t_M = T$ be a partitioning of the time interval $[0,T]$ with time steps $\Delta t_m := t_{m+1}-t_m$, $m=0, \ldots, M-1$. Smooth curves $\Gamma_i(t_m)$, $i=1, \ldots, N_C$, $m=0, \ldots, M$ are replaced by polygonal curves $\Gamma_i^m$ given by nodes $\vec X_{i,j}^m$ which are approximations of $\vec x_i(q_j^i,t_m)$. Further, let $\kappa_{i,j}^m$ be an approximation of $\kappa_i(q_j^i, t_m)$. The derivative terms with respect to time are replaced by difference quotients of the form
\begin{equation}
(\vec x_i)_t(q_i^j, t_m) \approx \frac{1}{\Delta t_m} \left(\vec X_{i,j}^{m+1}-\vec X_{i,j}^m\right). 
\end{equation}  
Let $h_{i,j-\frac12}^m = \vec X_{i,j}^m - \vec X_{i,j-1}^m$, $i=1,\ldots,N_C$, $j=1,\ldots, N_i$, be the distance between two neighboring nodes. For each curve, we define a discete normal vector field $\vec \nu_i^m$ by
\begin{equation*}
\vec\nu_i^m|_{[q_{j-1}^i, q_j^i]} := \vec\nu_{i,j-\frac12}^m :=\frac{(\vec X_{i,j}^m - \vec X_{i,j-1}^m)^\perp}{h_{i,j-\frac12}^m},
\end{equation*}
see also \cite{BGN07a,BGN07b, Benninghoff2014a}. Here, $\perp$ denotes the anti-clockwise rotation of a vector by $\pi/2$. Further, we define the following weighted approximating normal vector at
$\vec X_{i,j}^m$ by
\begin{equation*}
\vec\omega_{i,j}^m := \frac{h_{i,j-\frac12}^m \vec\nu_{i,j-\frac12}^m + h_{i,j+\frac12}^m \vec\nu_{i,j+\frac12}^m}{h_{i,j-\frac12}^m  + h_{i,j+\frac12}^m}
= \frac{(\vec X_{i,j+1}^m - \vec X_{i,j-1}^m)^\perp}{h_{i,j-\frac12}^m  + h_{i,j+\frac12}^m}, 
\end{equation*}
for $j=1,\ldots,N_i$ if $\partial \Gamma_i^m = \emptyset$ and for $j=1, \ldots, N_i-1$ if $\partial\Gamma_i^m \neq \emptyset$. 
In the latter case, we set 
\begin{align*}
\vec\omega_{i,0}^m &:= \vec\nu_{i,\frac12}^m = \frac{(\vec X_{i,1}^m-\vec X_{i,0}^m)^\perp}{h_{i,\frac12}^m},\\
\vec\omega_{i,N_i}^m &:= \vec\nu_{i,N_i-\frac12}^m = \frac{(\vec X_{i,N_i}^m-\vec X_{i,N_i-1}^m)^\perp}{h_{i,N_i-\frac12}^m}. 
\end{align*}
The external term $F$ is approximated by
\begin{align*}
F_{i,j}^m :=& \lambda \left[ ( u_0(\vec X_{i,j}^m) - u(\vec X_{i,j}^m + a\vec \omega_{i,j}^m) )^2 \right.\\
& \left.- ( u_0(\vec X_{i,j}^m) - u(\vec X_{i,j}^m - a\vec \omega_{i,j}^m) )^2 \right],
\end{align*}
with a small real number $a>0$ if $j$ is not the index of a free endpoint, and we set $F_{i,j}^m=0$ else. 

The equation for the normal velocity \eqref{eq:twodim_free_endpoints_eq_Vn} is approximated by
\begin{subequations}
\label{eq:discrete_scheme}
\begin{equation}
\frac{1}{\Delta t_m} \left(\vec X_{i,j}^{m+1}-\vec X_{i,j}^m\right) \,.\, \vec\omega_{i,j}^m = \sigma \kappa_{i,j}^{m+1} + F_{i,j}^m.
\label{eq:discrete_scheme1}
\end{equation}
Thus, for computing $\Gamma_i^{m+1}$, we use the previous curve $\Gamma_i^m$ for the external term $F_{i,j}^m$ and for the weighted normal $\vec\omega_{i,j}^m$. 

For an approximation of \eqref{eq:curvature_identity}, we need to define an approximation of $(\vec x_i)_{ss}(q_j^i, t_{m+1})$. For that, we make use of difference quotients of the form 
\begin{align*}
\Delta_2^{h,m} \vec X_{i,j}^{m+1} :=&  \frac{2}{h_{i,j-\frac12}^m  + h_{i,j+\frac12}^m} \left(  (\vec X_{i,j+1}^{m+1} - \vec X_{i,j}^{m+1})/h_{i,j+\frac12}^m \right. \\
& \left. - (\vec X_{i,j}^{m+1} - \vec X_{i,j-1}^{m+1})/h_{i,j-\frac12}^m\right),
\end{align*}
for $i=1, \ldots, N_C$ and $j=1, \ldots, N_i$, if $\partial \Gamma_i^m = \emptyset$, and $j=1, \ldots, N_i-1$, else. In case of equal spatial step sizes $h_{i,j-\frac12}^m = h_{i,j+\frac12}^m =: h_i^m$, the term reduces to $(\vec X_{i,j-1}^m - 2 \vec X_{i,j}^m + \vec X_{i,j+1}^m)/((h_i^m)^2)$, see also \cite{Benninghoff2014a}, where we also defined and used these difference quotients.

The equation \eqref{eq:curvature_identity} is now approximated by
\begin{equation}
\kappa_{i,j}^{m+1} \vec\omega_{i,j}^m = \Delta_2^{h,m} \vec X_{i,j}^{m+1},
\label{eq:discrete_scheme2}
\end{equation}
for $i=1,\ldots,N_C$, $j=1, \ldots, N_i$ in case of closed curves and $j=1, \ldots, N_i-1$ in case of open curves.

In case of open curves, additional equations for the endpoints are needed. The case of triple junctions and boundary intersection points is described in \cite{Benninghoff2014a}. For curves with free endpoints we introduce the tangential vectors $\vec\tau_{i,0}^m = (\vec X_{i,1}^m - \vec X_{i,0}^m)/h_{i,\frac12}$, and $\vec\tau_{i,N_i}^m = (\vec X_{i,N_i}^m - \vec X_{i,N_i-1}^m)/h_{i,N_i-\frac12}$. The equations \eqref{eq:Vtan0} and \eqref{eq:Vtan1} are approximated by
\begin{align}
&\frac{1}{\Delta t_m} \left(\vec X_{i,0}^{m+1}-\vec X_{i,0}^m\right) \,.\, \vec\tau_{i,0}^m = \nonumber\\
 =& \sigma - |\vec\tau_{i,0}^m\,.\,\vec e_1| (\nabla_h^2 u(\vec z^{0,1}))^2  -  |\vec\tau_{i,0}^m\,.\,\vec e_2| (\nabla_h^1 u(\vec z^{0,2}))^2,
\label{eq:discrete_scheme3}
\end{align}
and 
\begin{align}
&\frac{1}{\Delta t_m} \left(\vec X_{i,N_i}^{m+1}-\vec X_{i,N_i}^m\right) \,.\, \vec\tau_{i,N_i}^m = \nonumber\\
 =& -\sigma + |\vec\tau_{i,N_i}^m\,.\,\vec e_1| (\nabla_h^2 u(\vec z^{1,1}))^2  + |\vec\tau_{i,N_i}^m\,.\,\vec e_2| (\nabla_h^1 u(\vec z^{1,2}))^2.
\label{eq:discrete_scheme4}
\end{align}
\end{subequations}
Similarly, discrete versions of \eqref{eq:Vn0} and \eqref{eq:Vn1} can be stated using $\vec\omega_{i,0}^m$ and $\vec\omega_{i,N_i}^m$ as discrete normal vectors. 

The scheme \eqref{eq:discrete_scheme} is a numerical approximation of the scheme \eqref{eq:parametric_scheme_free_endpoints}, where the parametric curves are replaced by polygonal curves, and the smooth functions $\vec x_i$ and $\kappa_i$ are replaced by continuous functions uniquely given by their values at the nodes $q_j^i$, $i=1, \ldots, N_C$, $j=0, \ldots, N_i$. 

The discrete scheme can be rewritten to a linear system with a sparse system matrix, similar as presented in \cite{Benninghoff2014a}, and can be solved with a fast direct solver like for example the UMFPACK algorithm \cite{Davis04}. 

\subsection{Numerical Solution of the Denoising Problem}
For computing a numerical solution $u^h$ for the piecewise smooth, denoised version $u$ of $u_0$, we consider for $N_x, N_y \in \mathbb{N}$ the discrete set
\begin{equation*}
\Omega^h := \left\{(ih,jh)\,:\, i=0, \ldots, N_x, \,j=0\ldots,N_y\right\},
\end{equation*}
where $N_x$ and $N_y$ are  the number of pixels in $x$- and $y$-direction. 
We define for $i=1,\ldots, N_x$, $j=1,\ldots, N_y$
\begin{align*}
A_x(i,j) &= \left\{
\begin{array}{ll}
h^2, & \text{if }\quad [(i-1)h, ih]\times \{j\} \cap \Gamma_{i_0}^m = \emptyset, \\
     & \forall i_0 \in \{1,\ldots, N_C\},\\
0,   & \text{else,}
\end{array}
\right. 
\end{align*}
\begin{align*}
A_y(i,j) &= \left\{
\begin{array}{ll}
h^2, & \text{if }\quad \{i\}\times [(j-1)h, jh] \cap \Gamma_{i_0}^m = \emptyset, \\
     & \forall i_0 \in \{1,\ldots, N_C\},\\
0,   & \text{else.}
\end{array}
\right. \\
\end{align*}

Fixing the set of curves $\Gamma$, we consider the following discrete energy:
\begin{align}
E_{\mathrm{discr}}(u^h) 
=&   \sum_{i=1}^{N_x} \sum_{j=1}^{N_y} \left( A_x(i,j) \left(\frac{u_{i,j}^h - u_{i-1,j}^h}{h}\right)^2 \right.\nonumber\\
 & + \left. A_y(i,j) \left(\frac{u_{i,j}^h - u_{i,j-1}^h}{h}\right)^2 \right)\nonumber\\
 & + \lambda \sum_{i=0}^{N_x} \sum_{j=0}^{N_y} h^2 \left( u_0(ih,jh) - u_{i,j}^h\right)^2, 
\label{eq:denoising_energy}
\end{align}
which is a discrete analogue of $\int_{\Omega\setminus \Gamma} \left(\|\nabla u\|^2 \, \mathrm{d}x + \int_\Omega \lambda (u_0-u)^2 \right) \,\mathrm{d}x$. Here $u_{i,j}^h$ approximates $u$ at the node $(ih,jh)$. The piecewise continuous function $u^h$ is uniquely given by its value at the points in $\Omega^h$. 

By setting the terms $A_x(i,j)$ or $A_y(i,j)$ to zero at points where the line $[(i-1)h,jh), (ih,jh)]$ or $[(ih,(j-1)h), (ih,jh)]$ intersects with one of the curves, we approximate the integral over the set $\Omega \setminus \Gamma$.  

Taking the derivative of the right hand side of \eqref{eq:denoising_energy}  with respect to $u_{i,j}^h$ and setting the resulting term to zero, leads to a linear system. The corresponding system matrix is sparse since each node $(ih,jh)\in \Omega^h$ is only coupled to a few neighboring nodes. The resulting linear system can be solved with a fast direct or iterative solver by employing the sparse matrix structure. 

Considering $h \rightarrow 0$, we obtain in the limit $\nabla u \,.\, \vec\nu = 0$ at the curves belonging to $\Gamma$, and $\nabla u \,.\, \vec n_{\partial\Omega} = 0$ at the image boundary $\partial \Omega$, where $\vec n_{\partial\Omega}$ is a normal vector field at $\partial \Omega$. For details, we refer to \cite{Benninghoff2014a}. Consequently, we obtain an edge preserving image smoothing if $\Gamma$ matches with the edges in the given image. 

\subsection{Topology Changes}
\label{subsec:topology_changes}
During the evolution of curves, topology changes can occur, since the edge set in the image and the boundaries of objects are not known in advance. Therefore, curves can split into two or more subcurves, curves can merge to one single curve, triple junctions and new curves may occur and curves can intersect with the image boundary such that new boundary nodes emerge. Further, a curve needs to be deleted if its length becomes too small. In \cite{Benninghoff2014a}, we extended the idea of \cite{Balazovjech12, MikulaUrban12}, and described a method to detect topology changes of curves efficiently. The main idea is the use of an artificial background grid which covers the entire image domain $\Omega$. We consider successively all nodes $\vec X_{i,j}^m$ and mark a grid element with $(i,j)$ if $\vec X_{i,j}^m$ is the first node located in this array. If a grid element is already marked with $(i_1,j_1)$ and the nodes $\vec X_{i,j}^m$ and $\vec X_{i_1,j_1}^m$ are not neighbor nodes, a topology change likely occurs close to this pair. Details on this method for curves without free endpoints are given in \cite{Benninghoff2014a}.  

In principle, topology changes involving curves with free endpoints can be detected similarly by using such a background grid. In addition to the topology changes listed above (splitting, merging, emergence of triple junctions and boundary intersection points), topology changes involving the free endpoints can occur: If two free endpoints of one curve are located in one square of the background grid, an open contour becomes a closed contour. If two free endpoints of two different curves meet, the two curves merge to one single curve, and the former free endpoints become inner nodes of the new curve. If a free endpoint and an inner point of a curve meet, a triple junction is created.

\subsection{Summary of the Algorithm}
We propose the following algorithm for image segmentation and image restoration with parametric contours with possible free endpoints:

Given a set of polygonal curves $\Gamma^0 = (\Gamma_1^0, \ldots, \Gamma_{N_C}^0)$ and $\vec X^0 = (\vec X_1^0, \ldots, \vec X_{N_C}^0)$ with $\vec X_i^0([0,1])=\Gamma_i^0$, perform the following steps for $m=0, 1, \ldots, M-1$: 
\begin{enumerate}
\item \label{step1} Compute a denoised image approximation $u^h$ by minimizing \eqref{eq:denoising_energy} (solve the corresponding sparse linear system).
\item \label{step2} Compute the external terms $F_{i,j}^m$ by using the solution $u^h$ of step \ref{step1}. Compute $\vec X^{m+1}$ by solving the linear equation derived from the scheme \eqref{eq:discrete_scheme}.
\item \label{step3} Check whether topology changes occur. If so, execute the topology change. 
\end{enumerate}
A segmentation of the image is given by the final set of curves $\Gamma^M$. An image restoration is given by the image approximation $u^h$ from the time step $t_M$.

\subsection{Modifications}
Step \ref{step2} of the algorithm above can additionally be split in two sub-steps: First, we fix the free endpoints and we let the inner nodes of the curve evolve. Then, we let the endpoints evolve according to the above presented discrete scheme. 

The main effort of this method compared to the Chan-Vese method for interface curves is that we have to solve a two-dimensional diffusion equation (bulk equation) several times during the segmentation. In the experiments described in the next section, we perform 10 steps of curve evolution followed by a solution of the bulk equation.  Having computed $u^h$, we use it for the next 10 curve evolution steps. 


As an alternative, we can start the segmentation using interface curves and the image segmentation method described in \cite{Benninghoff2014a} (based on the Chan-Vese method \cite{Chan01}) with piecewise constant approximations. As a postprocessing step, we can consider the derivatives of the image function in normal direction at the final curves (or the jump of the image function across the curves). We replace interface curves by curves with free endpoints if the derivatives in normal direction are locally very small. For that, we delete those parts of a curve where the derivative is small which results in curves with free endpoints. Next, we compute some steps of the segmentation method with free endpoints to obtain the final contours. 

Topology changes occur only in rare cases when using a postprocessing evolution of curves with free endpoints. In most situations, topology changes are already detected in the previous evolution.

\section{Results and Discussion}
\label{sec:results}
The method for image segmentation and restoration presented in sections \ref{sec:im_proc_methods} and \ref{sec:numerics} are applied on some exemplary test images. For all experiments presented in this section, we use constant time steps sizes $\Delta t_m = \Delta t$, $m=0,\ldots,M-1$.

\begin{figure}[t]
	\centering
		\includegraphics[viewport = 175 300 420 550, width=0.15\textwidth]{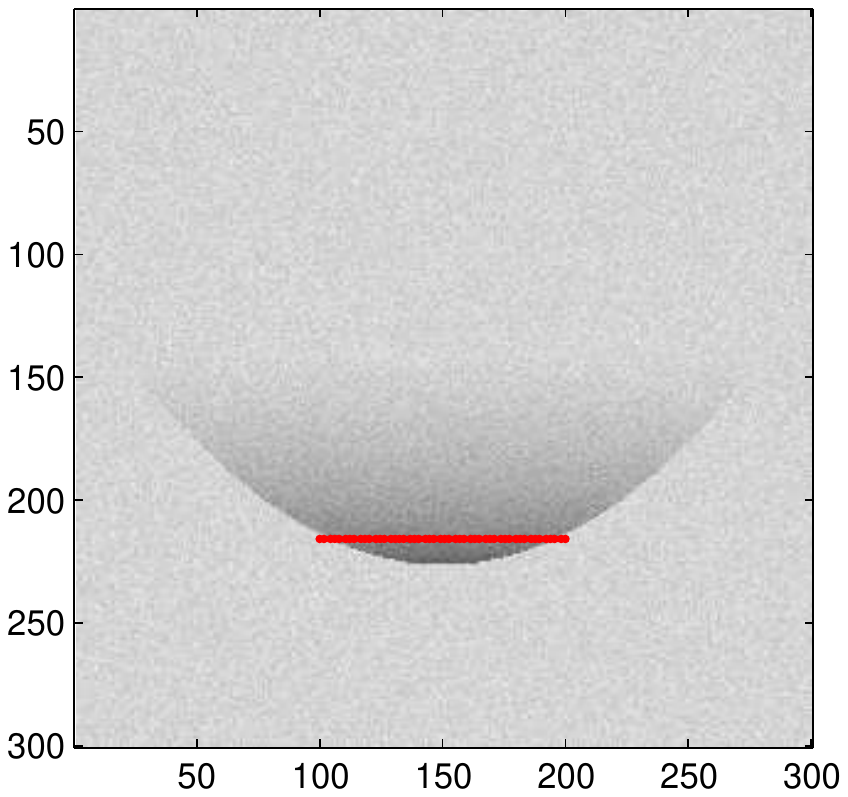}
		\includegraphics[viewport = 175 300 420 550, width=0.15\textwidth]{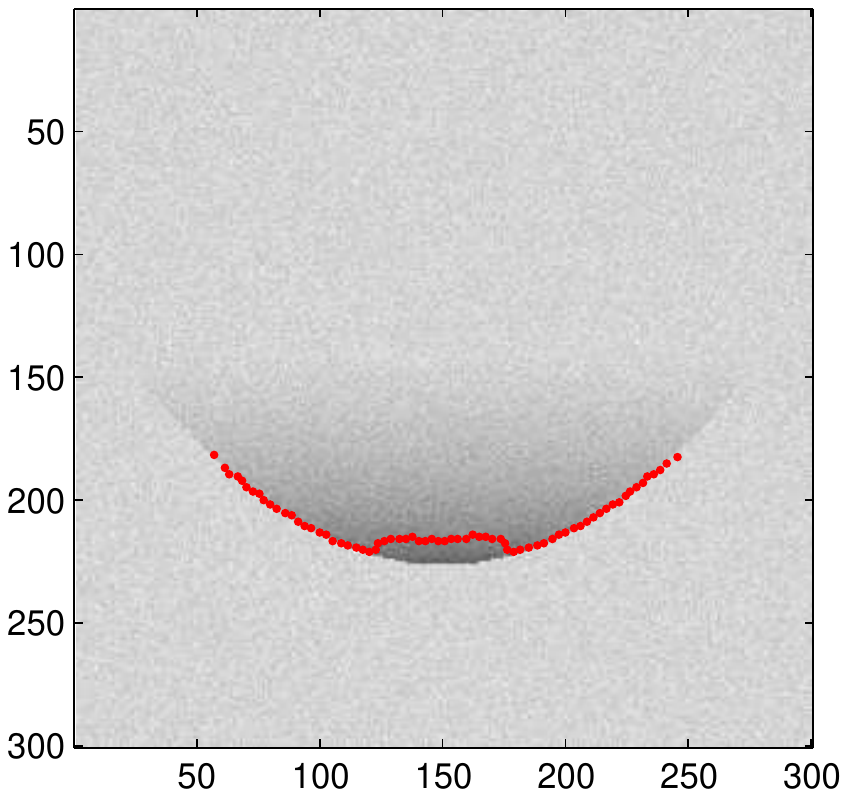}
		\includegraphics[viewport = 175 300 420 550, width=0.15\textwidth]{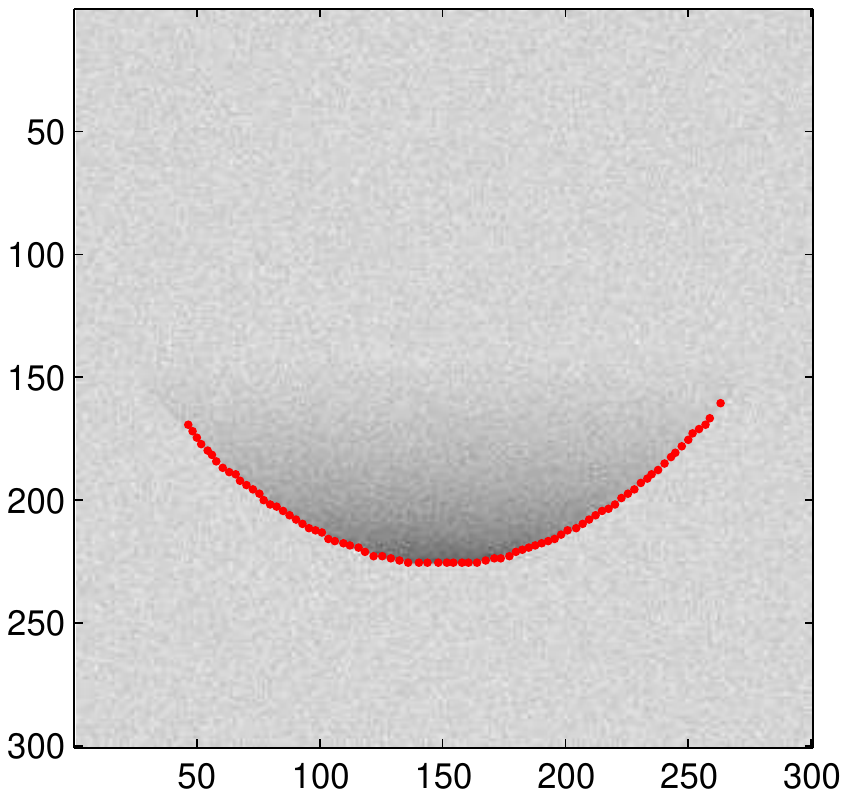}\\
		\includegraphics[viewport = 175 300 420 550, width=0.15\textwidth]{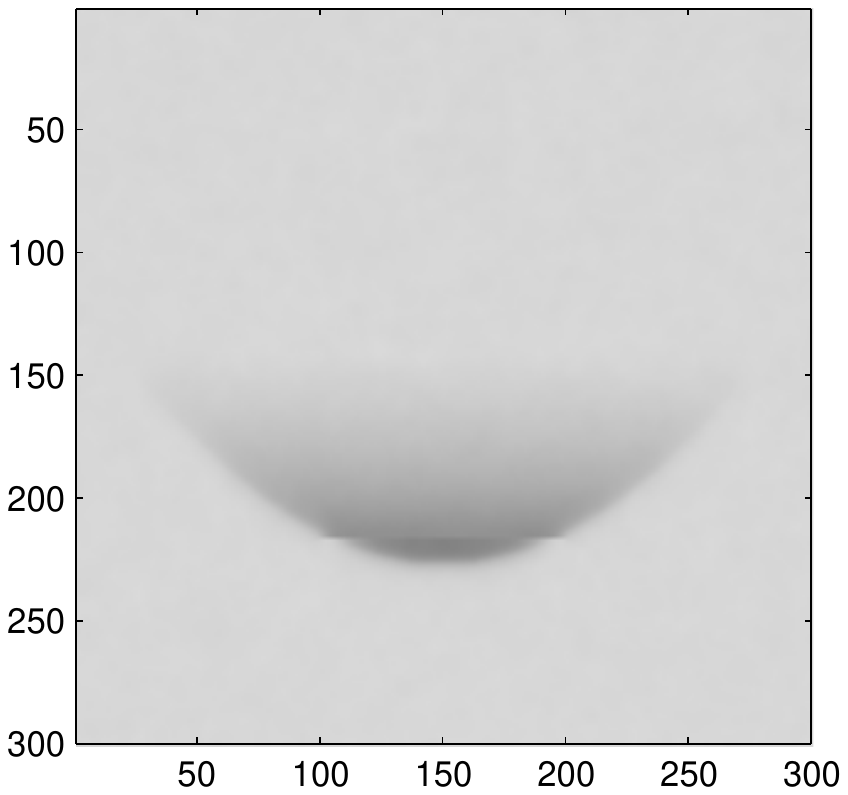}
		\includegraphics[viewport = 175 300 420 550, width=0.15\textwidth]{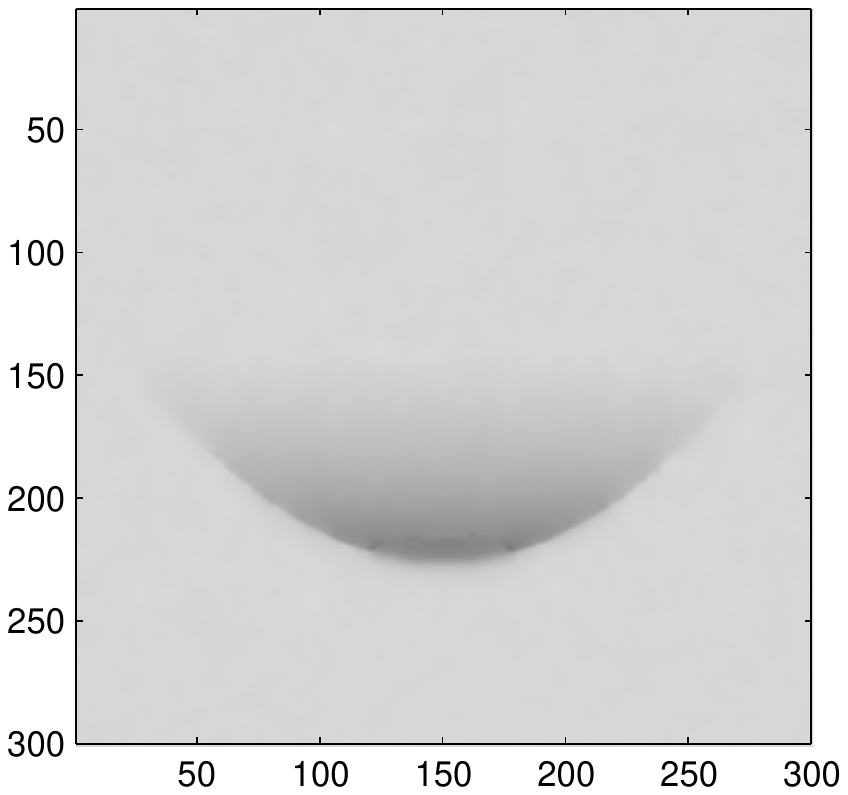}
		\includegraphics[viewport = 175 300 420 550, width=0.15\textwidth]{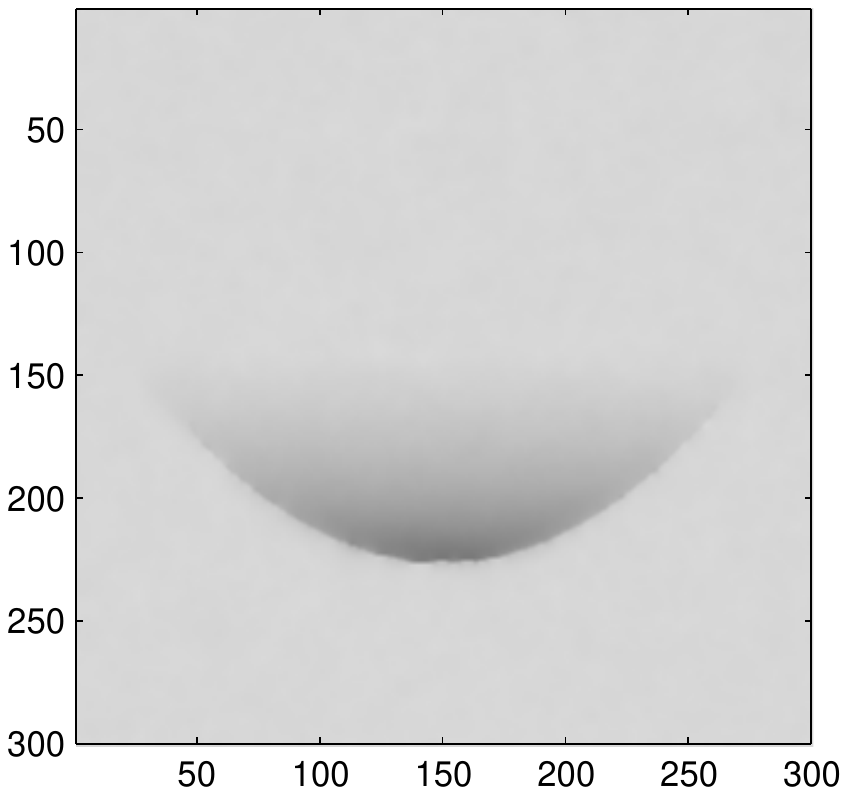}
	\caption{Example image showing a contour with two free endpoints. Original image and evolving contours (1st row) and denoised image (2nd row) for $m=1,1000,6000$ using $\Delta t=0.032$, $\sigma = 2\mathrm{e}-5$ and $\lambda = 0.002$.}
	\label{fig:result_free_endpoints}
\end{figure}

In the first experiment, we consider an example where a contour with two free endpoints evolves in the image domain and detects an edge. 
Figure~\ref{fig:result_free_endpoints}  presents the results of image segmentation and denoising. It can be observed that the image is not smoothed out across the curve $\Gamma$. Further a growth of the curve in tangential direction can be observed. The growth stops when the inequalities \eqref{eq:free_endpoints_inequality1} and \eqref{eq:free_endpoints_inequality2} become equalities. This depends on the absolute values of the difference quotients $|\nabla_h^i u|$, $i=1,2$, and the weighting parameter $\sigma$. The image approximation $u$ attends values in $[0,1]$. In this image,  differences of the form $u(\vec x + h \vec e_i) - u(\vec x)$ are typically of magnitude $10^{-2}$. Since $\Omega = [1,300] \times [1,300]$ and $h=1$, $|\nabla_h^i u|^2$ is of magnitude $10^{-4}$. Therefore, we have to choose a small value for the weighting parameter $\sigma$, here, we choose $\sigma = 2\mathrm{e}-5$. If we used a normalized image domain $\Omega = [0,1] \times [0,1]$, the pixel grid would have a grid size of $h=1/300$ and $h^2 = 1/90000$. In this case, we would choose a weight $\sigma$ of magnitude $1$.   

In a second experiment, we study a crack tip problem which has also been considered in \cite{Pock2009}. The image function is given by 
\begin{equation}
u_0(\vec x) = a\sqrt{r(\vec x)} \sin(\theta(\vec x)/2) + b, 
\end{equation}
where $r(\vec x)\geq 0$, $\theta(\vec x) \in (-\pi,\pi]$ are polar coordinates with $r=0$ corresponding to the image center, and $a,b\in \mathbb{R}$ are constants such that $u_0$ attends values in $[0,1]$. 

\begin{figure}[t]
	\centering
		\includegraphics[viewport = 175 300 420 550, width=0.15\textwidth]{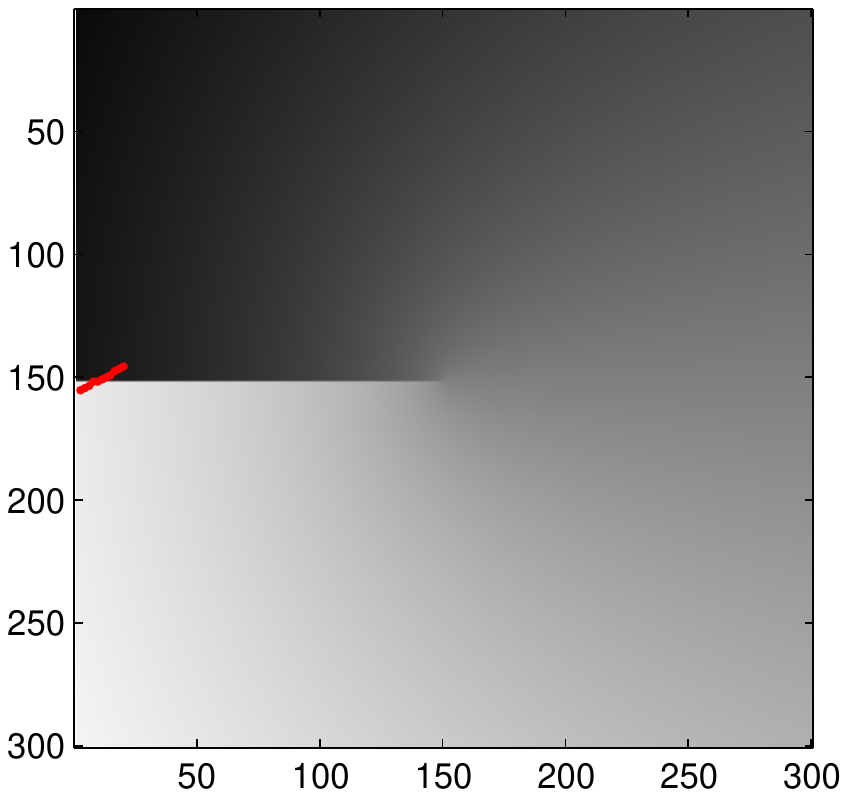}
		\includegraphics[viewport = 175 300 420 550, width=0.15\textwidth]{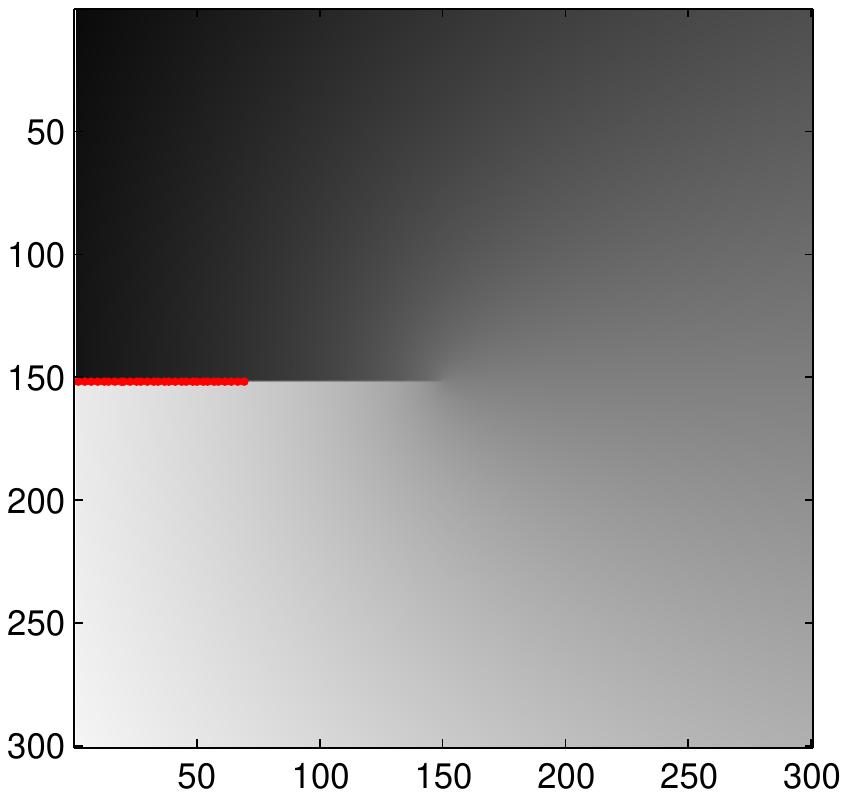}
		\includegraphics[viewport = 175 300 420 550, width=0.15\textwidth]{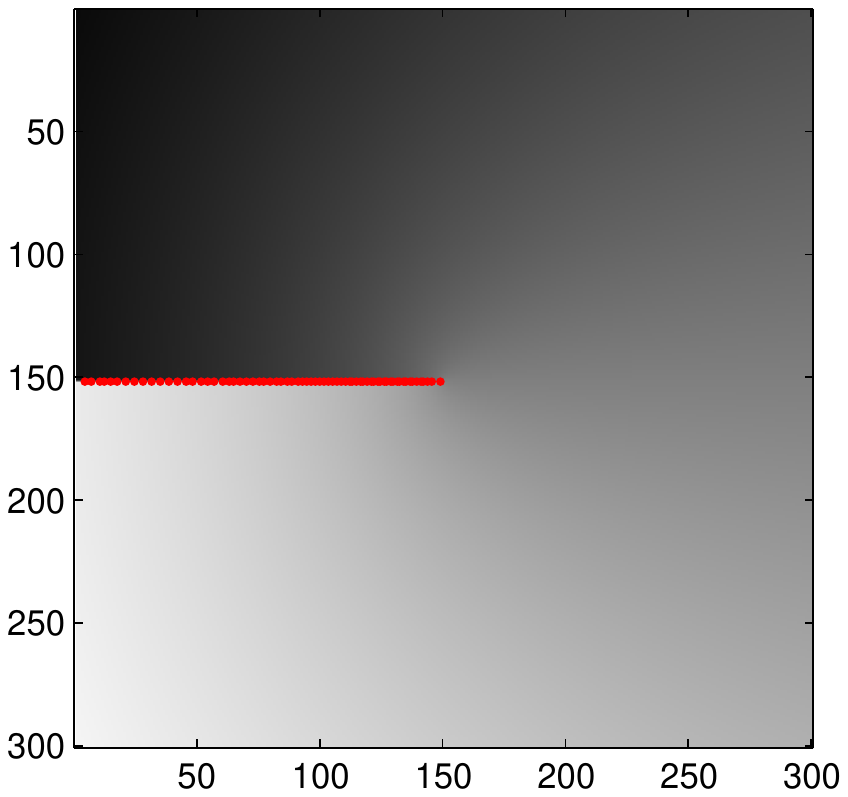}
	\caption{Example image showing a contour with one free endpoint and one boundary intersection point, see also \cite{Pock2009}. Evolving contours for $m=1,500,3000$ using $\Delta t=0.001$, $\sigma = 2\mathrm{e}-5$ and $\lambda = 0.002$.}
	\label{fig:result_free_endpoints3_pock}
\end{figure}

Figure~\ref{fig:result_free_endpoints3_pock} shows the evolution of a contour with one free endpoint. The second endpoint belongs to the image boundary. At time step $m=3000$, the free endpoint is located at the image center and the curve matches with the edge in the image. As discussed above (see also Equations \eqref{eq:free_endpoints_inequality1} and \eqref{eq:free_endpoints_inequality2}), the parameter $\sigma$, which weights the length term, needs to be chosen small enough such that the curve can extend. If $\sigma$ is fixed, the absolute value of the difference quotients  must be large enough such that the length of the curve increases. In this example, the edge is a horizontal line and the position where the curve stops depends on the value of $\nabla_h^2 u$, i.e. on the difference quotient in $y$-direction. 

\begin{figure}[t]
	\centering
		\includegraphics[viewport = 175 300 420 550, width=0.15\textwidth]{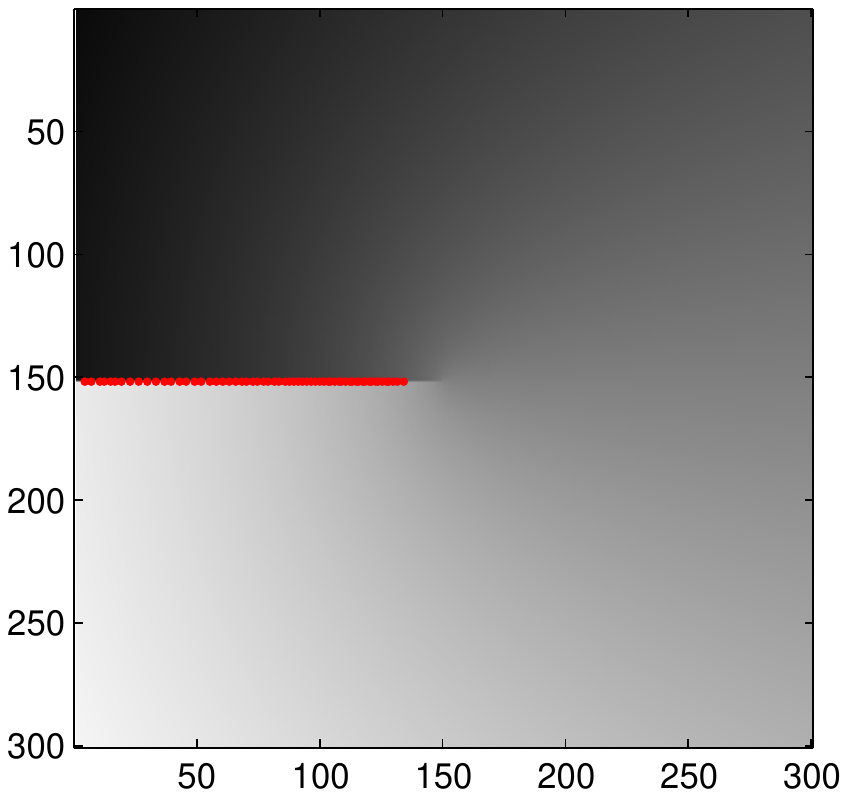}
		\includegraphics[viewport = 175 300 420 550, width=0.15\textwidth]{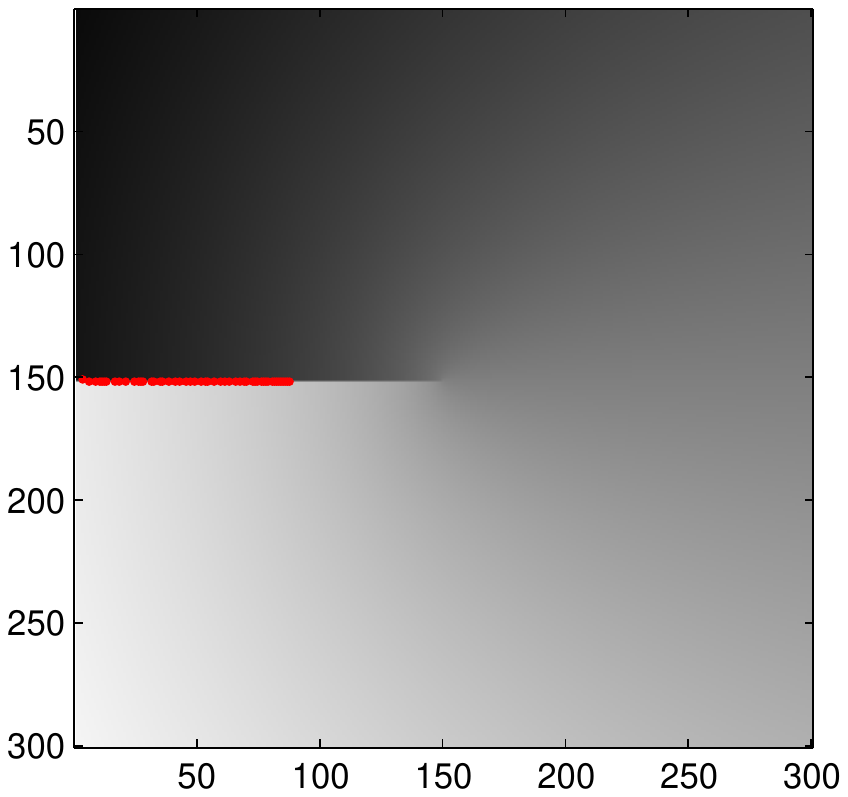}
	\caption{Dependency on the weighting parameter using $\sigma = 0.002$ (left) and $\sigma = 0.01$ (right), $\Delta t=0.001$ and $\lambda = 0.002$ for $m=3000$. If a too large weight is chosen for the length term in \eqref{eq:mumford_shah_discrete}, the curve does not reach the image center.}
	\label{fig:result_free_endpoints3_pock_largersigma}
\end{figure}

We rerun the example using $\sigma=0.002$ and $\sigma=0.01$ instead of $\sigma=2\mathrm{e}-5$. Figure~\ref{fig:result_free_endpoints3_pock_largersigma} shows the results at time step $m=3000$. In both cases, the free endpoint does not reach the center of the image since the value of $\sigma$ has been set larger. The growth of the curve already stops at larger values of $\nabla_h^2 u$, recall conditions \eqref{eq:free_endpoints_inequality1} and \eqref{eq:free_endpoints_inequality2}). We even let the curve evolve until time step $m=5000$, but we observed no significant motion between $m=3000$ and $m=5000$.

\begin{figure}[t]
	\centering
		\includegraphics[viewport = 175 300 420 550, width=0.15\textwidth]{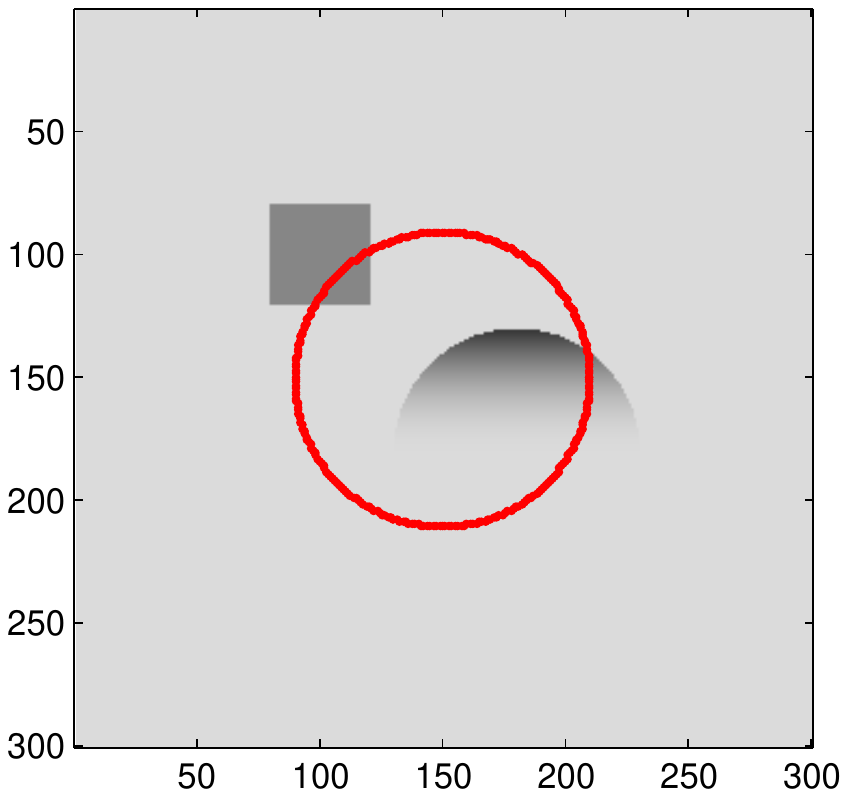}
		\includegraphics[viewport = 175 300 420 550, width=0.15\textwidth]{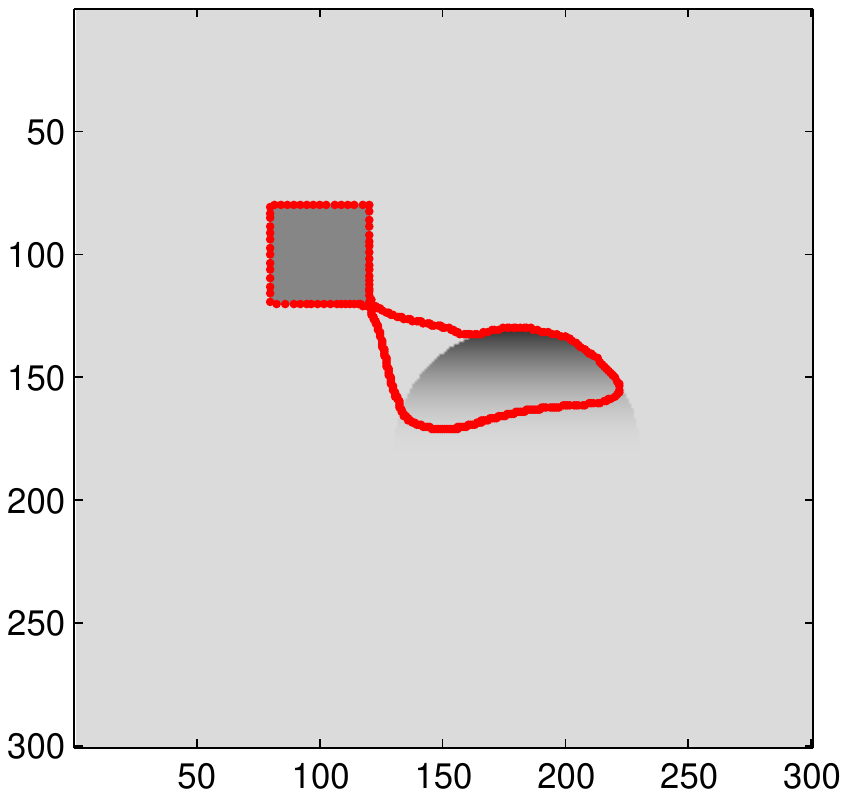}
		\includegraphics[viewport = 175 300 420 550, width=0.15\textwidth]{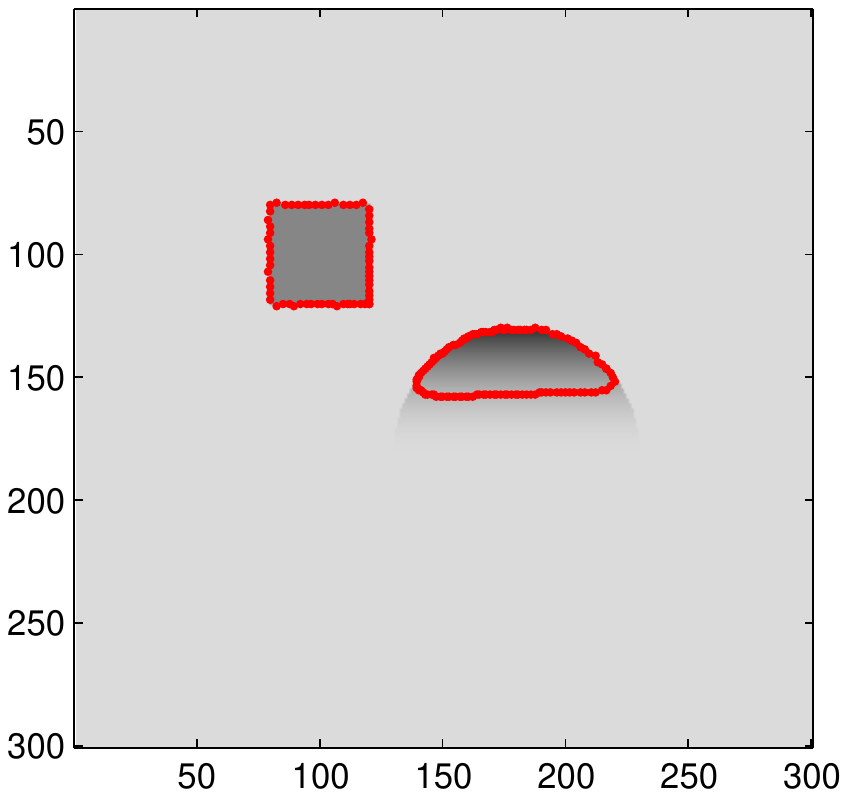}\\
		\includegraphics[viewport = 175 300 420 550, width=0.15\textwidth]{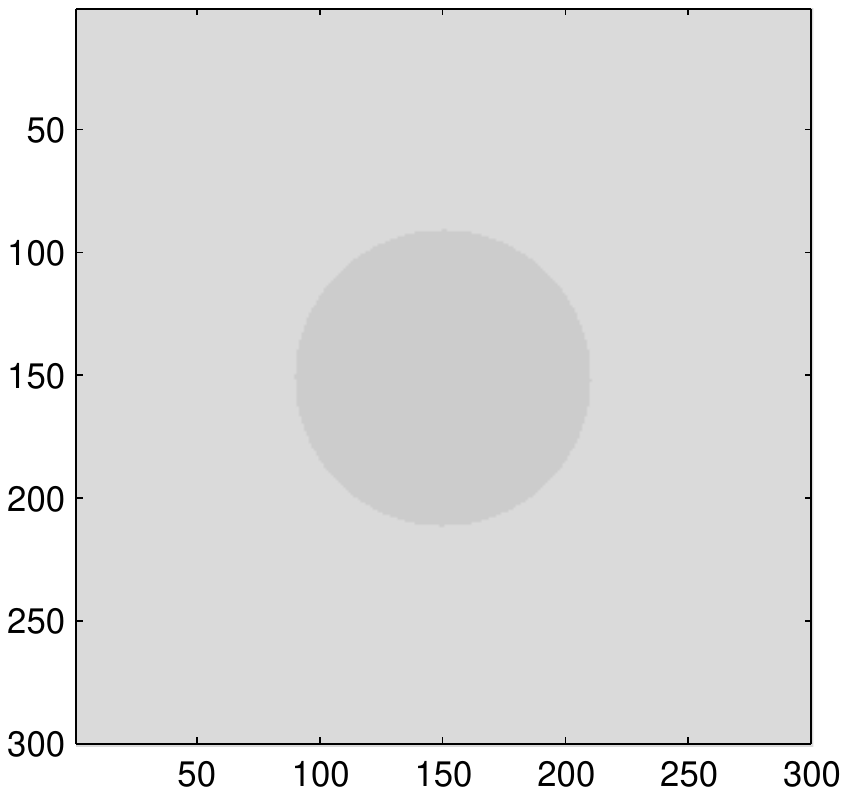}
		\includegraphics[viewport = 175 300 420 550, width=0.15\textwidth]{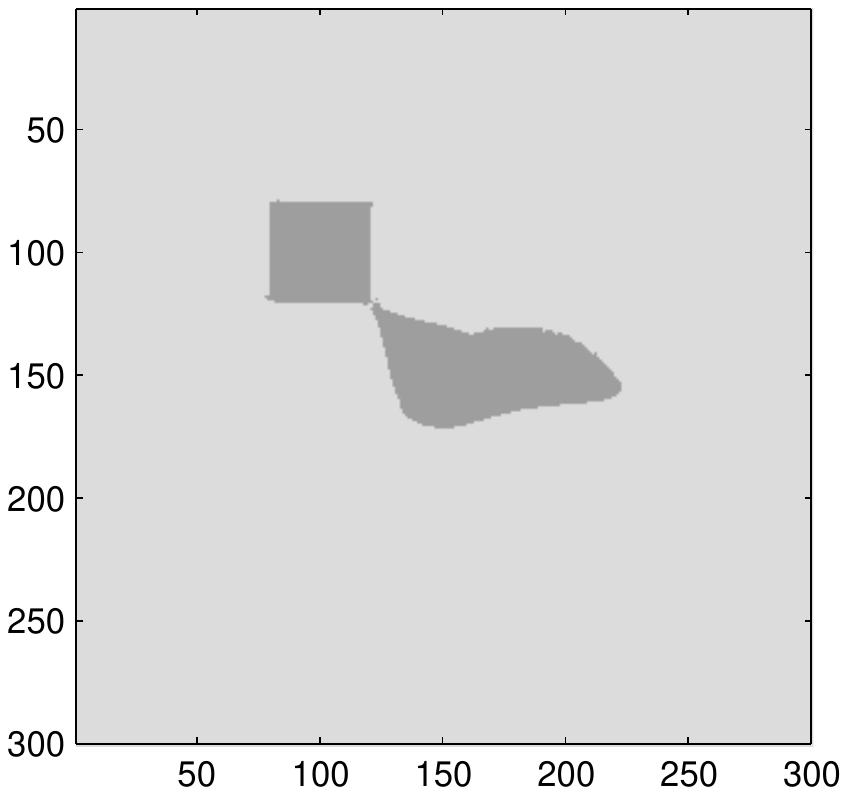}
		\includegraphics[viewport = 175 300 420 550, width=0.15\textwidth]{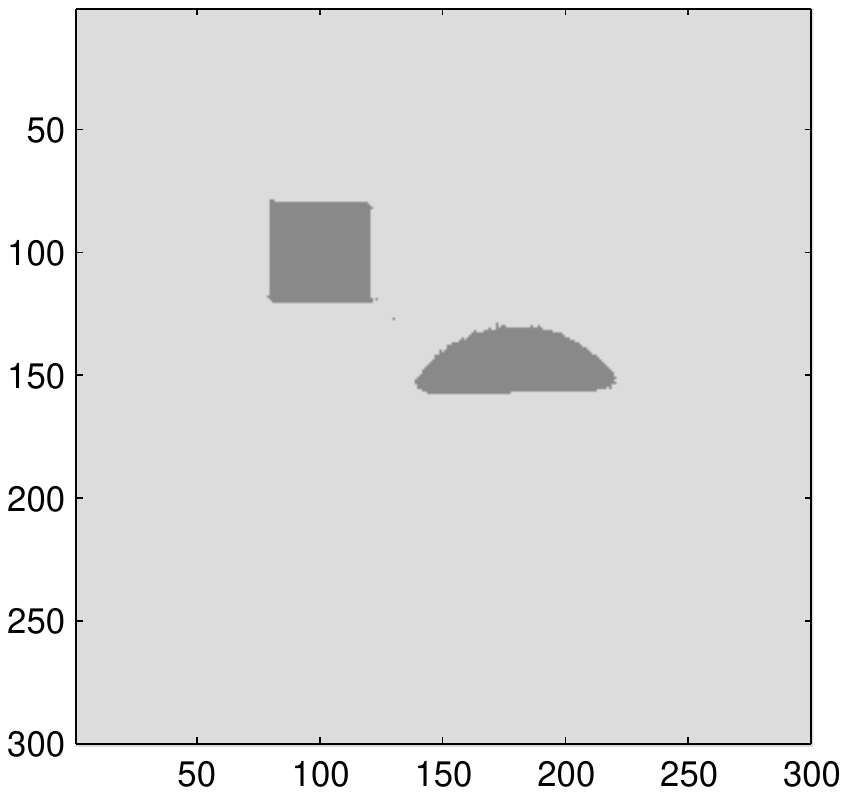}
	\caption{Image segmentation result using Chan-Vese, interface curves and a piecewise constant image approximation. Original image and contours (first row) and piecewise constant approximation (second row) for $m=1, 1050, 1200$.}
	\label{fig:result_free_endpoints2_pre}
\end{figure}

In another experiment, which demonstrates the evolution of curves with free endpoints, we first apply the parametric method of \cite{Benninghoff2014a} to the Chan-Vese problem \cite{Chan01} using interface-curves. This means, that we first segment a given image in regions separated by interface curves, see Figure~\ref{fig:result_free_endpoints2_pre}. We start with one large initial curve which splits up in two sub-curves. This example thus also demonstrates the handling of a topology change.

\begin{figure}[t]
	\centering
		\includegraphics[viewport = 175 300 420 550, width=0.15\textwidth]{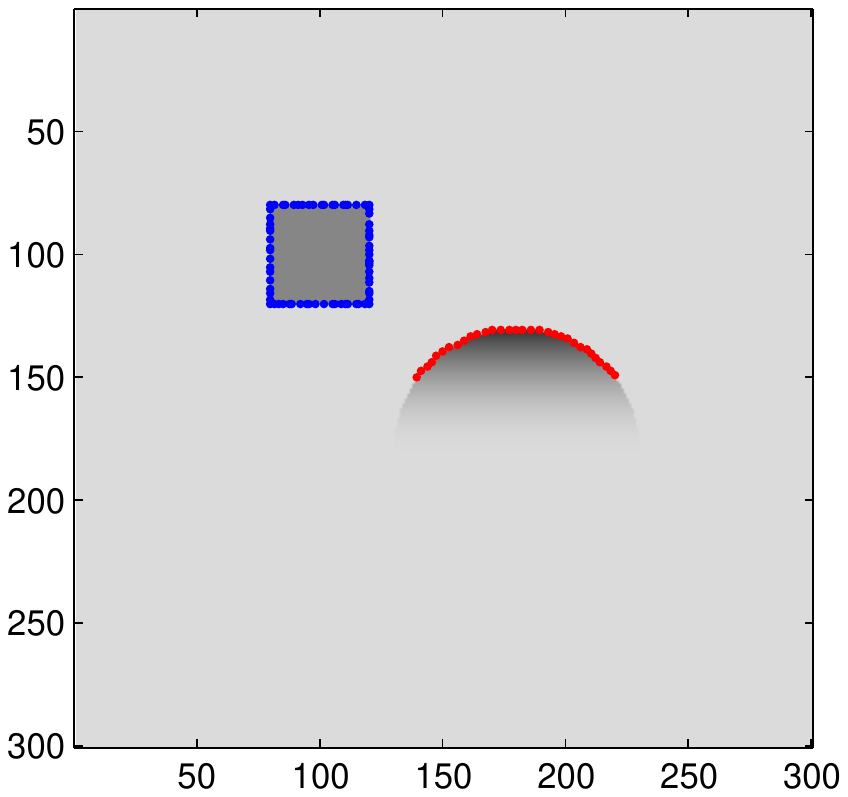}
		\includegraphics[viewport = 175 300 420 550, width=0.15\textwidth]{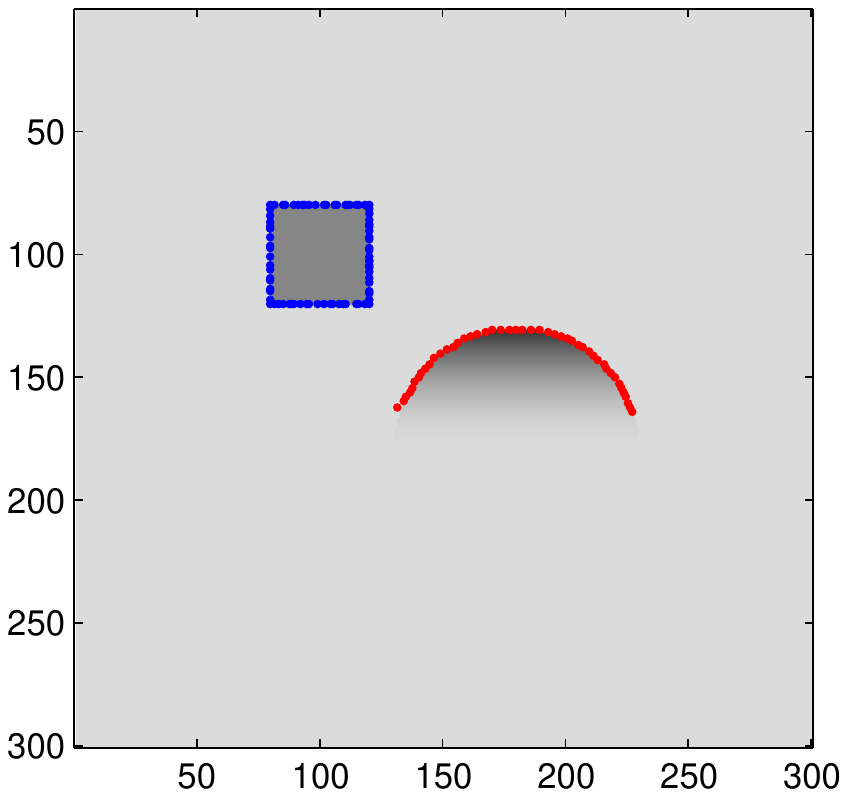}
		\includegraphics[viewport = 175 300 420 550, width=0.15\textwidth]{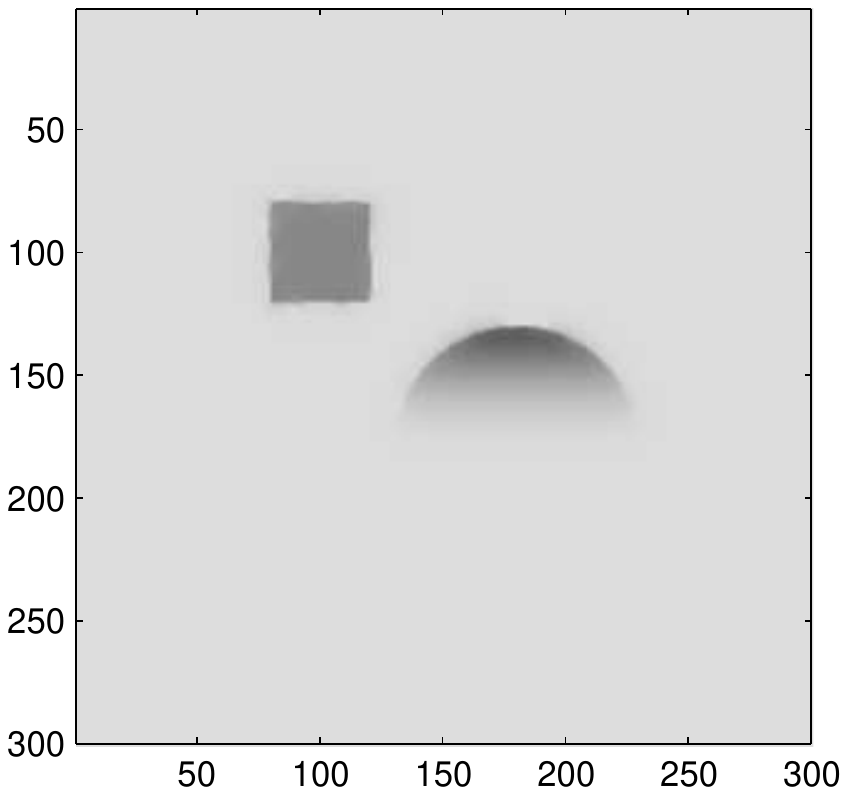}
	\caption{Postprocessing evolution with a contour with two free endpoints using $tol=0.1$ to obtain the initial contour. Original image and contours (left, center) for $m=1, 400$ using $\Delta t = 0.05$, $\sigma=2\mathrm{e}-5$,  $\lambda = 2\mathrm{e}-4$ and denoised image for $m=400$ (right).}
	\label{fig:result_free_endpoints2_post}
\end{figure}

In a postprocessing step, we delete those nodes where the jump of $u_0$ across the curve is smaller than a given tolerance of $tol=0.1$. This results in one closed curve, where no points are deleted (blue curve in Figure~\ref{fig:result_free_endpoints2_post}), and in one curve with two free endpoints (red curve). Figure~\ref{fig:result_free_endpoints2_post} shows the results of a postprocessing evolution of the curve. This example shows that our methods for image segmentation and denoising can be applied also on images with both open and closed edges. 

\begin{table}[!t]
\renewcommand{\arraystretch}{1.3}
\caption{Comparison of discrete Mumford-Shah Energy}
\label{tab:MS_energy_free_endpoints}
\centering
\begin{tabular}{p{0.13\textwidth}p{0.08\textwidth}p{0.1\textwidth}}
\toprule
Processing Method & Step Nr & Discrete Mumford-Shah Energy \\  
\midrule
Chan-Vese, piecewise constant & 1200 (final) & 22364.94\\
Postprocessing, free endpoints & 1 (start) &  23162.04\\
Postprocessing, free endpoints & 400 (final) &  18284.36\\
\bottomrule
\end{tabular}
\end{table}

Table~\ref{tab:MS_energy_free_endpoints} shows the values of the discrete Mumford-Shah energy \eqref{eq:mumford_shah_discrete} for the last step of the Chan-Vese piecewise constant segmentation with closed region boundaries (cf. Figure~\ref{fig:result_free_endpoints2_pre}, $m=1200$) and for the initial and final step of the postprocessing with one open boundary (cp. Figure~\ref{fig:result_free_endpoints2_post}, $m=1$ and $m=400$). Note, that the absolute values are large, since the image consists of $90000$ pixels and the size of each pixel is $1 \times 1$. The average contribution of each pixel to the energy is $<1$; however, it sums up to a value of magnitude $2 \cdot 10^4$.

\begin{figure}[t]
	\centering
		\includegraphics[viewport = 175 300 420 550, width=0.15\textwidth]{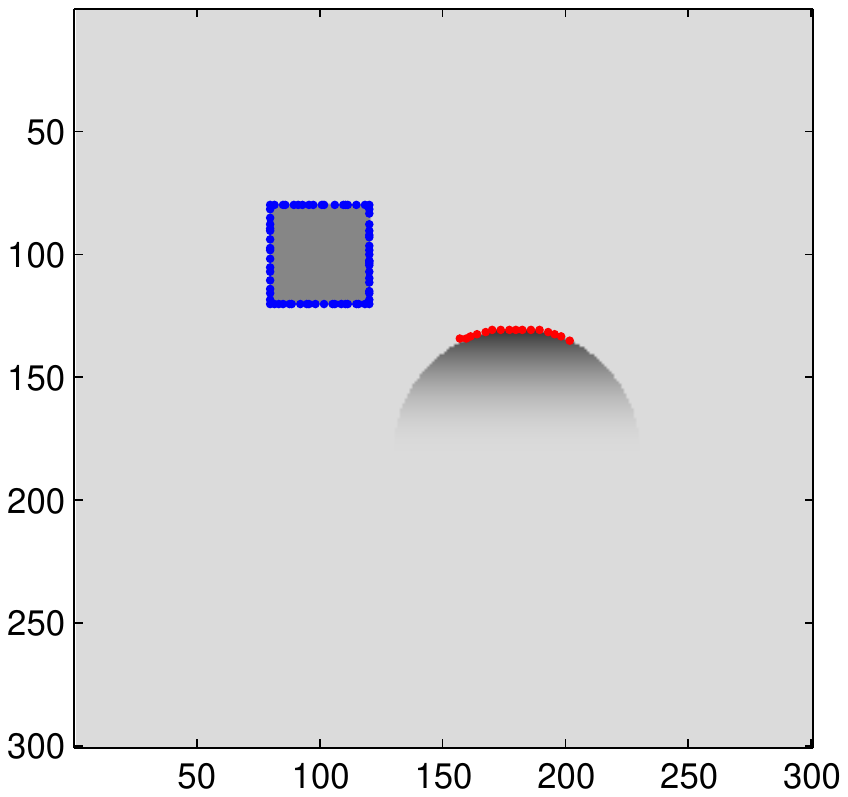}
		\includegraphics[viewport = 175 300 420 550, width=0.15\textwidth]{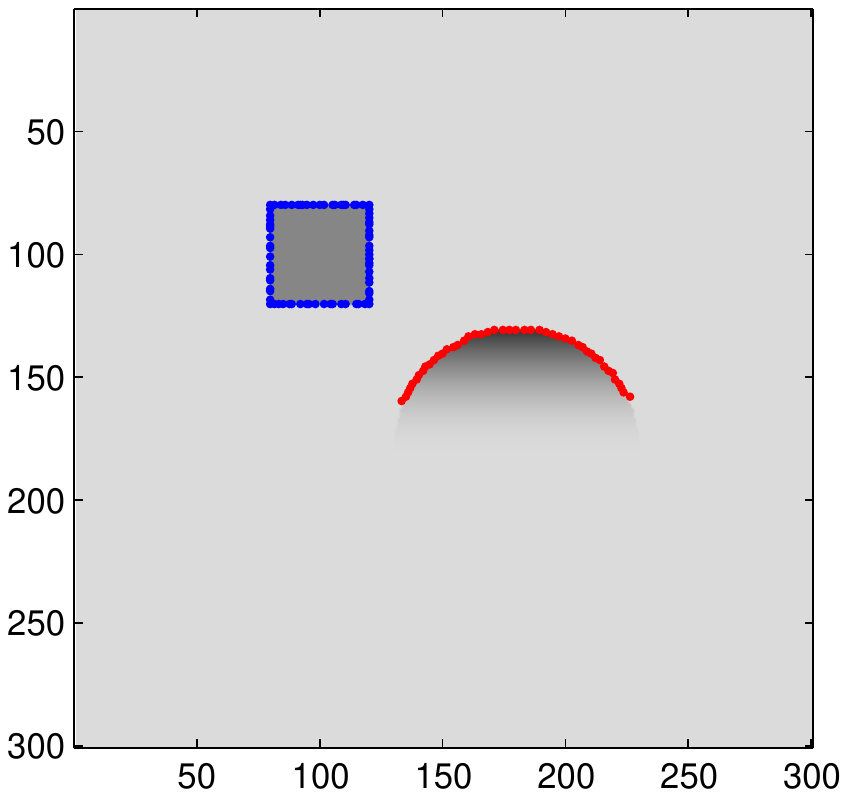}
		\includegraphics[viewport = 175 300 420 550, width=0.15\textwidth]{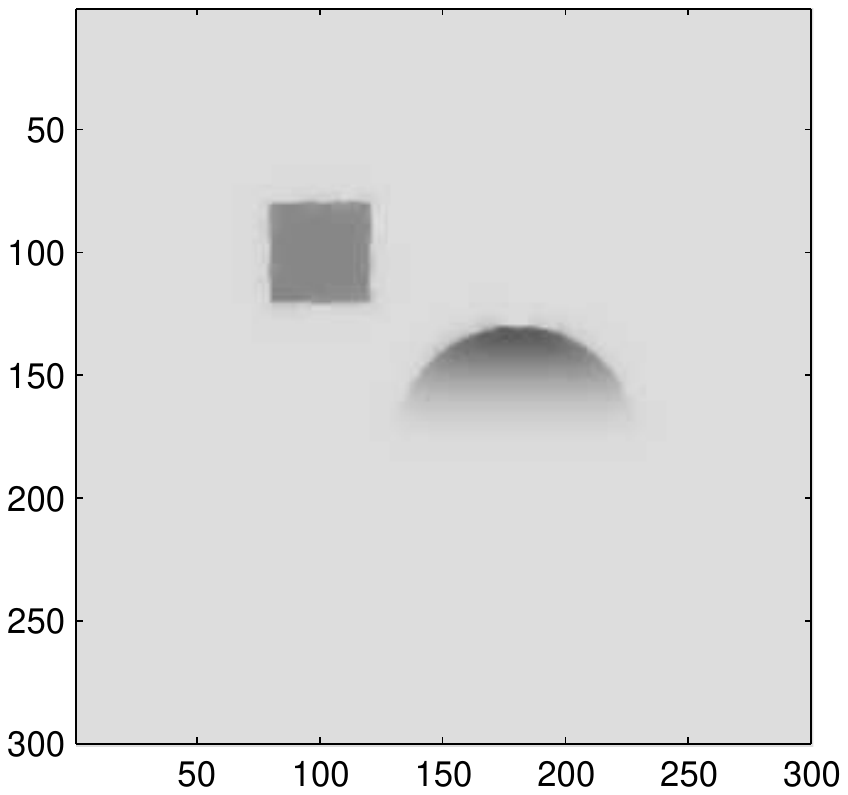}
	\caption{Postprocessing evolution with a contour with two free endpoints using $tol=0.5$ to obtain the initial contour. Original image and contours (left, center) for $m=1, 600$ using $\Delta t = 0.05$, $\sigma=2\mathrm{e}-5$,  $\lambda = 2\mathrm{e}-4$ and denoised image for $m=600$ (right).}
	\label{fig:result_free_endpoints2_post2}
\end{figure}

From Table~\ref{tab:MS_energy_free_endpoints}, we can observe that the energy even slightly increases from the last step of the Chan-Vese piecewise constant method to the first step of the postprocessing evolution. Deleting part of the curve does \textit{not} decrease the energy in this example. However, at the end of the postprocessing evolution with one open contour, we obtain a decrease of the energy. The energy at $m=400$ of the postprocessing is $81.75\%$ of the energy of the last step of the piecewise constant segmentation with closed boundaries. Therefore, if we delete part of the curve and if we let the curve with free endpoints evolve again, we will obtain a final curve such that the corresponding discrete Mumford-Shah energy  \eqref{eq:mumford_shah_discrete} is reduced compared to the Chan-Vese piecewise constant result. 

Next, we investigate the influence of the tolerance value $tol$, which is used for the deletion of some nodes of the curve. Note, that the image function attends values in $[0,1]$ where $0$ corresponds to black and $1$ corresponds to white color. The exact value of the tolerance $tol$ influences only the start curve of the second curve evolution. We repeat the postprocessing evolution and use $tol=0.5$ as tolerance resulting in a different initial curve. 

Figure~\ref{fig:result_free_endpoints2_post2} shows the postprocessing evolution of the curve. Of course, since the initial contour in Figure~\ref{fig:result_free_endpoints2_post2} (left) is smaller compared to the initial contour in Figure~\ref{fig:result_free_endpoints2_post}, more iteration steps are needed to obtain the final contour. 

The final result is independent on the exact initial curve as long as it is of the same type, i.e. open with free endpoints; not closed or not fully deleted. 
In our example the largest jump of $u_0$ across the final red curve of the first evolution (cf. Figure~\ref{fig:result_free_endpoints2_pre}, right sub-figures) is $0.61$, the smallest jump is $0.05$. The large difference between the largest and smallest jump can be used as an indicator to replace the interface-curve by a curve with two free endpoints. (On the contrary, the jump across the blue curve is constant in this example.) As tolerance value $tol$ any value larger than $0.05$ and smaller than $0.61$ could be chosen. For $tol\leq 0.05$ no node point would be deleted resulting in an unchanged curve. For $tol\geq 0.61$ all nodes and therefore the entire curve would be deleted. All values between the two thresholds can be theoretically used. Therefore, in this example, the final result is independent on the exact value of $tol$ as long it is in $(0.05, 0.61)$.

\begin{figure}[t]
	\centering
		\includegraphics[viewport = 220 300 370 540, width=0.1\textwidth]{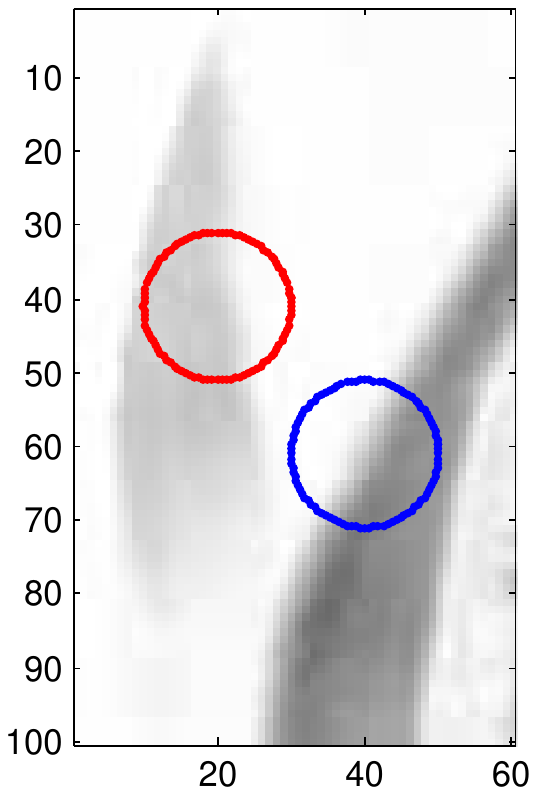}
		\includegraphics[viewport = 220 300 370 540, width=0.1\textwidth]{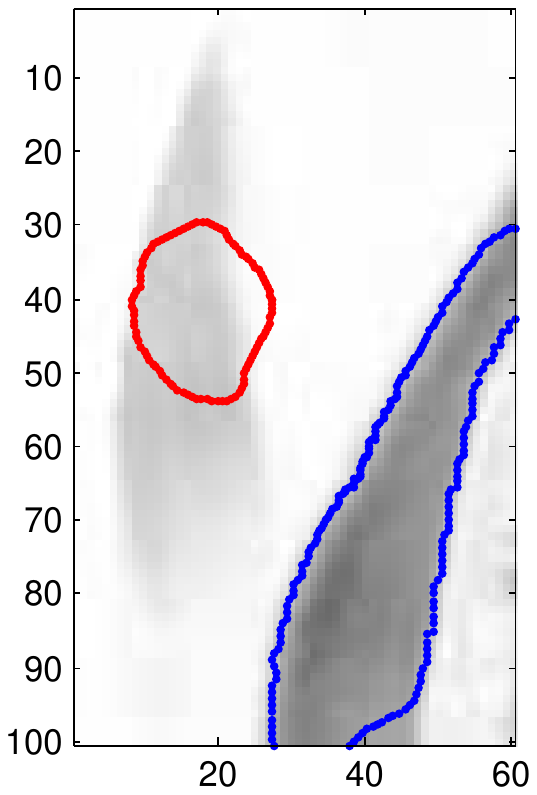}
		\includegraphics[viewport = 220 300 370 540, width=0.1\textwidth]{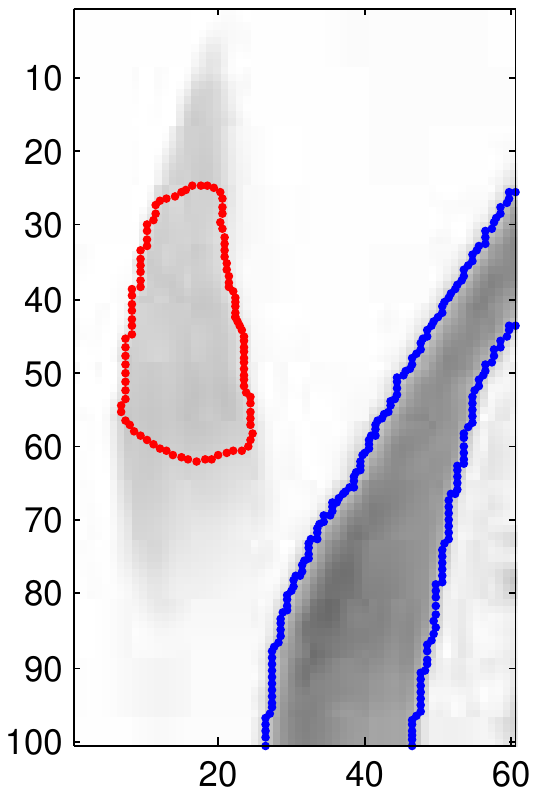}
		\includegraphics[viewport = 220 300 370 540, width=0.1\textwidth]{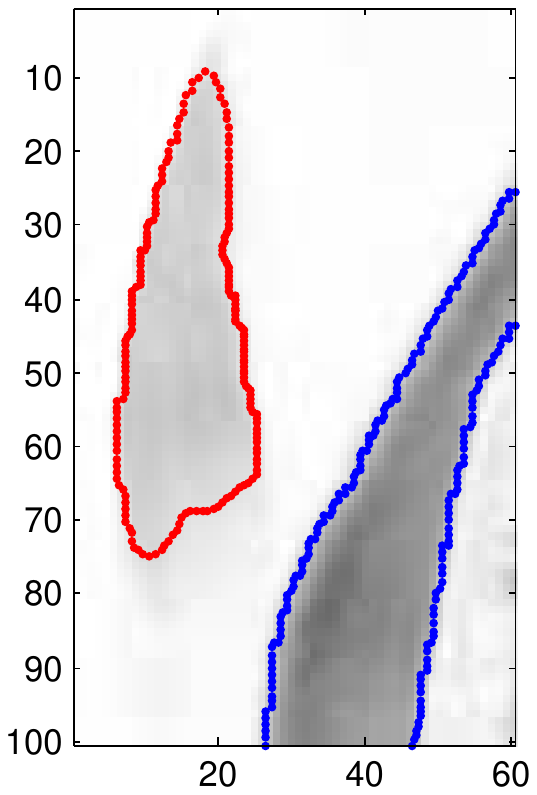}
	\caption{Medical image segmentation using Chan-Vese, interface curves and a piecewise constant image approximation. Original image and contours for $m=1, 400, 1000, 2500$. Image courtesy: Dr. Declan O'Regan and the Robert Steiner MR Unit, MRC Clinical Sciences Centre, Imperial College London.}
	\label{fig:result_medical_pre}
\end{figure}

\begin{figure}[t]
	\centering
		\includegraphics[viewport = 220 300 370 540, width=0.1\textwidth]{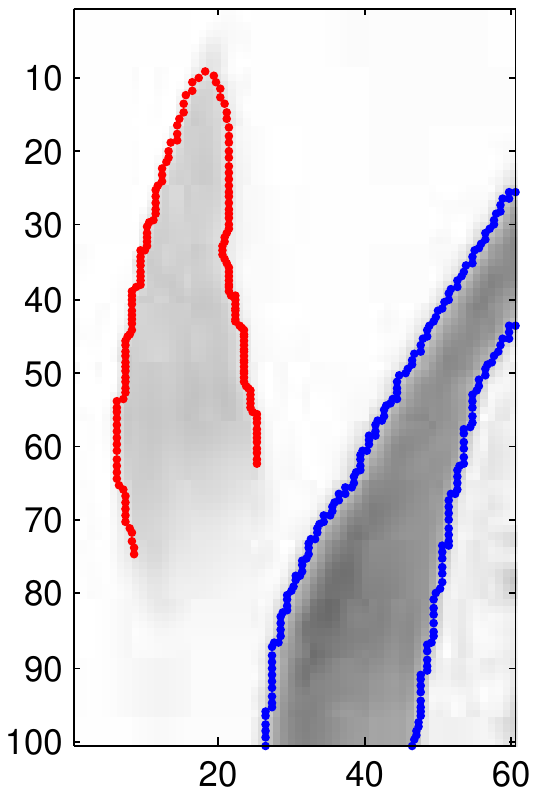}
		\includegraphics[viewport = 220 300 370 540, width=0.1\textwidth]{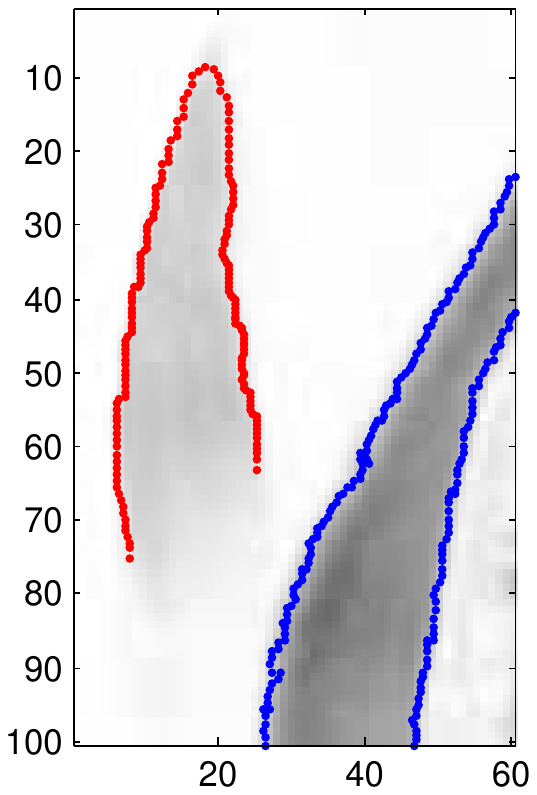}
		\includegraphics[viewport = 220 300 370 540, width=0.1\textwidth]{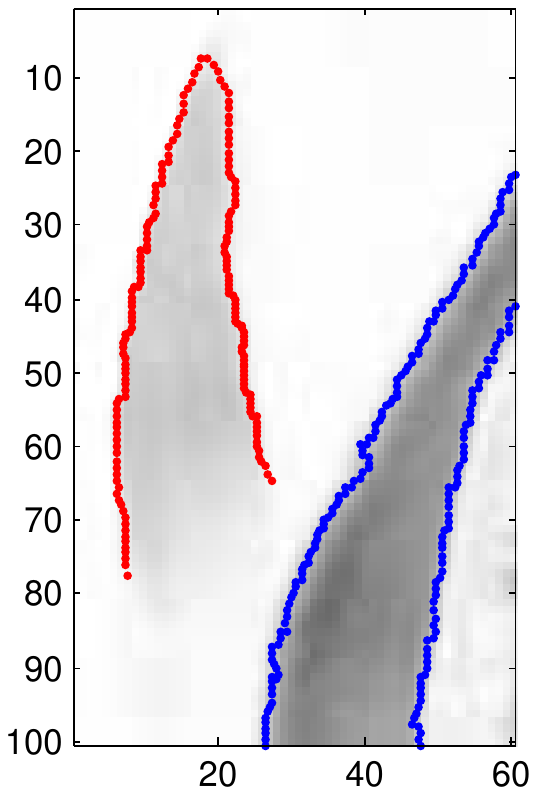}
		\includegraphics[viewport = 220 300 370 540, width=0.1\textwidth]{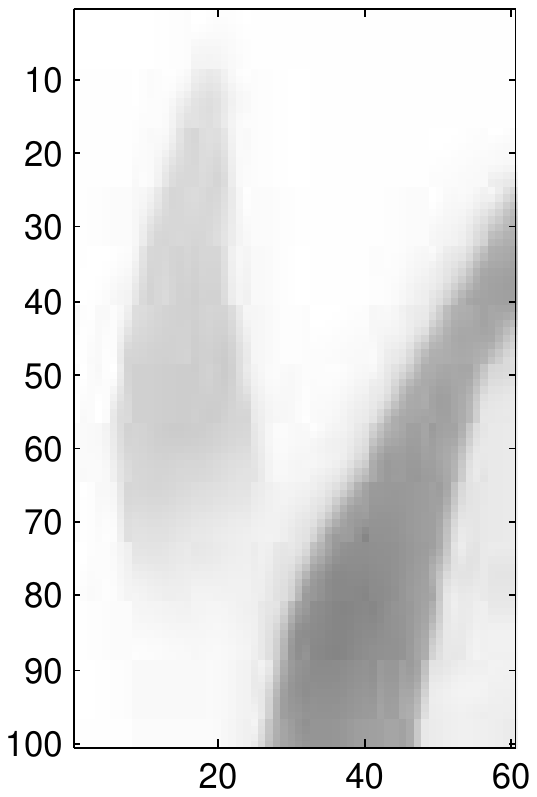}
	\caption{Postprocessing segmentation with free endpoints.  Original image and contours (far left, left, right) for  $m=1, 100, 500$ using $\Delta t = 0.008$, $\sigma=3.3\mathrm{e}-5$,  $\lambda = 0.0167$ and denoised image (far right) for $m=500$. Image courtesy: Dr. Declan O'Regan and the Robert Steiner MR Unit, MRC Clinical Sciences Centre, Imperial College London.}
	\label{fig:result_medical_post}
\end{figure}

Next, we demonstrate an example where a real, medical image is processed. Figure~\ref{fig:result_medical_pre} and Figure~\ref{fig:result_medical_post} show an excerpt of a medical image and the result of an edge detection. We first use the Chan-Vese algorithm with piecewise constant image approximation for segmenting the image, see Figure~\ref{fig:result_medical_pre}. In this first segmentation step, also topology changes occur. The initial closed curve touches twice the image boundary and splits up in two open curves each with two boundary intersection points. After the preceding segmentation, a part of the red curve is deleted (using a tolerance of $0.1$ for the jump across the curve) resulting in a curve with free endpoints. Figure~\ref{fig:result_medical_post} shows the result of the postprocessing evolution. Small tangential motions of the free endpoints can be observed. 

\begin{figure}[t]
	\centering
		\includegraphics[viewport = 150 280 440 560, width=0.15\textwidth]{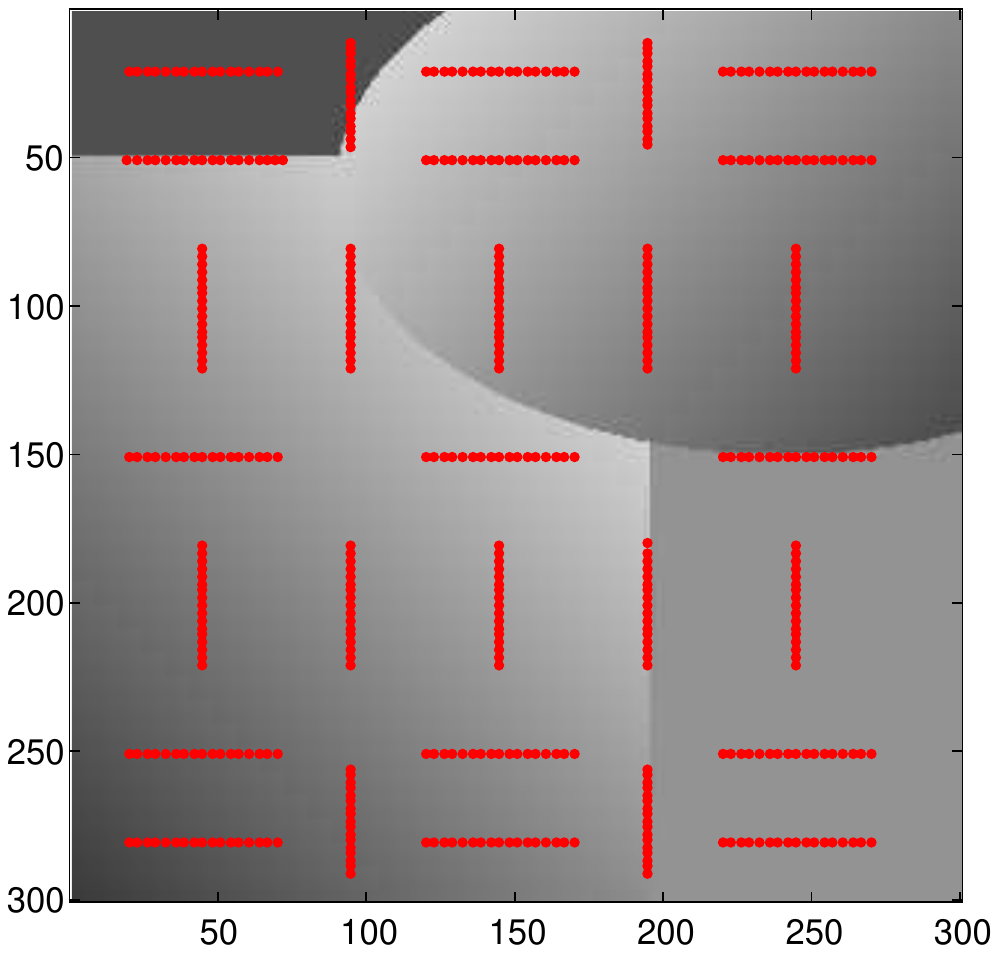}
		\includegraphics[viewport = 150 280 440 560, width=0.15\textwidth]{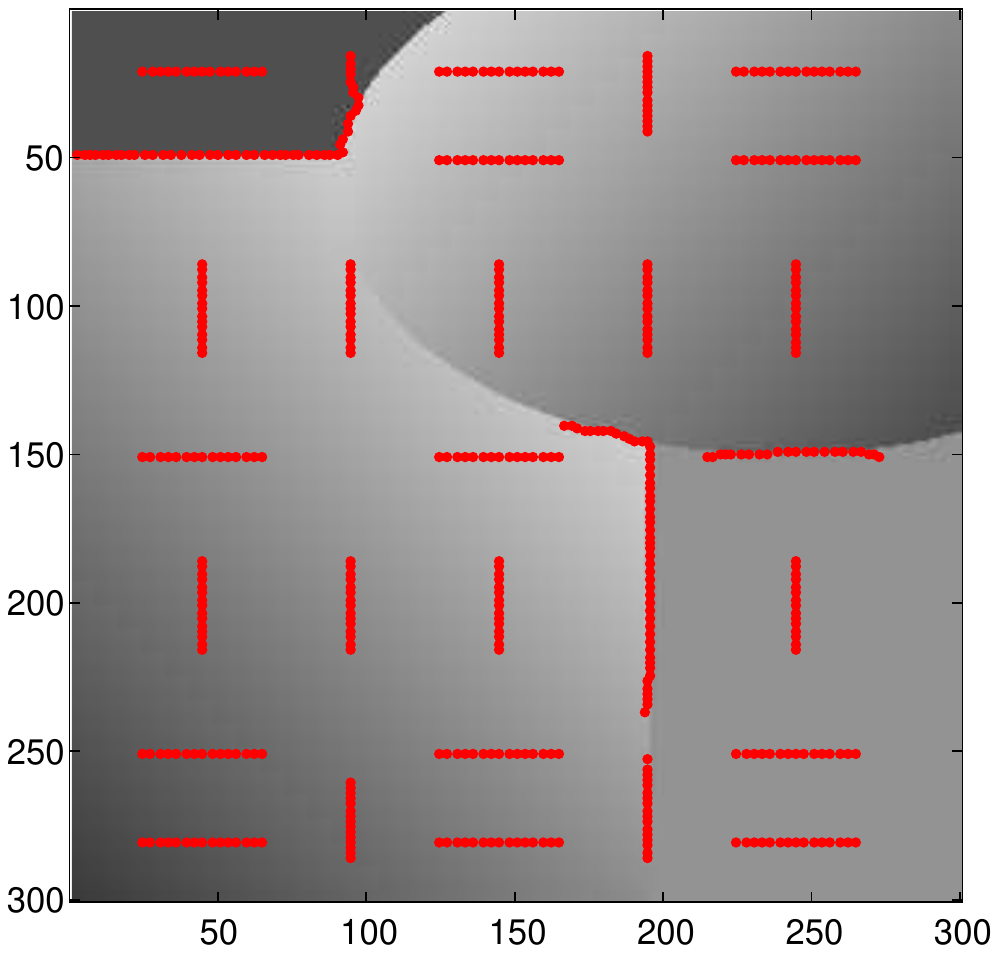}
		\includegraphics[viewport = 150 280 440 560, width=0.15\textwidth]{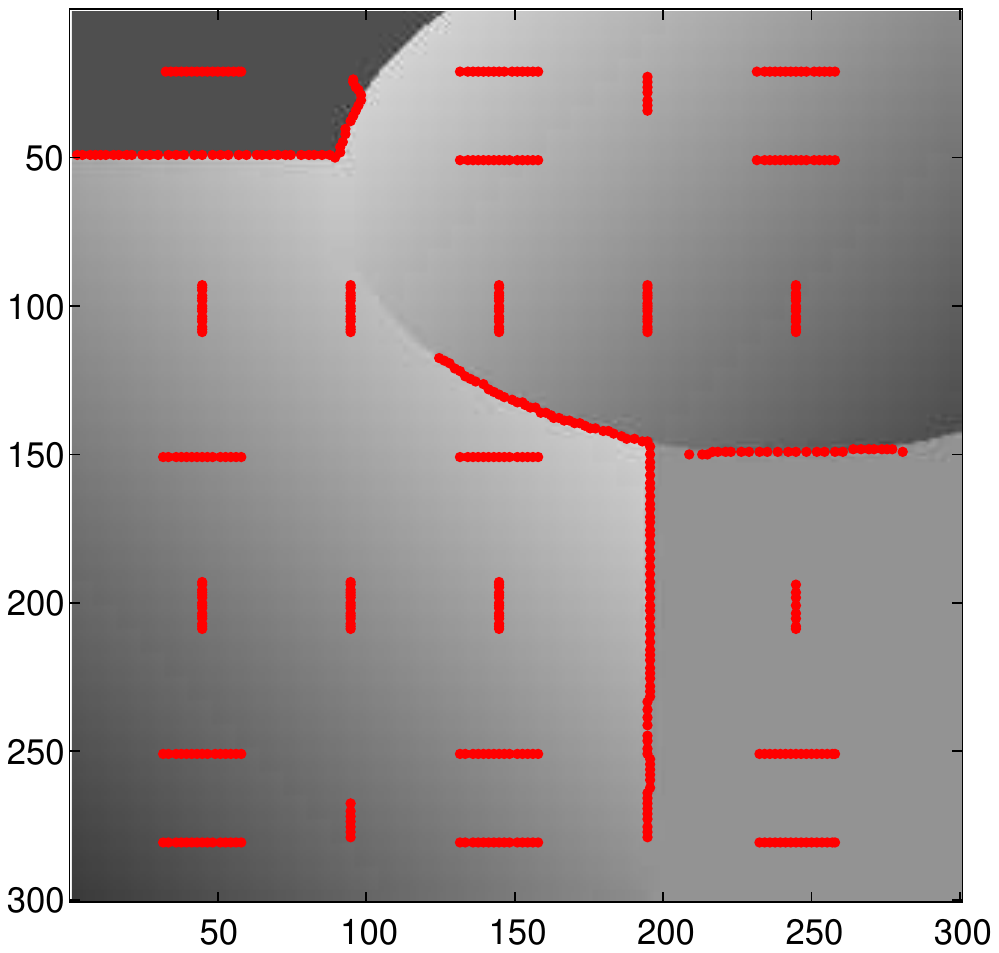}\\
		\includegraphics[viewport = 150 280 440 560, width=0.15\textwidth]{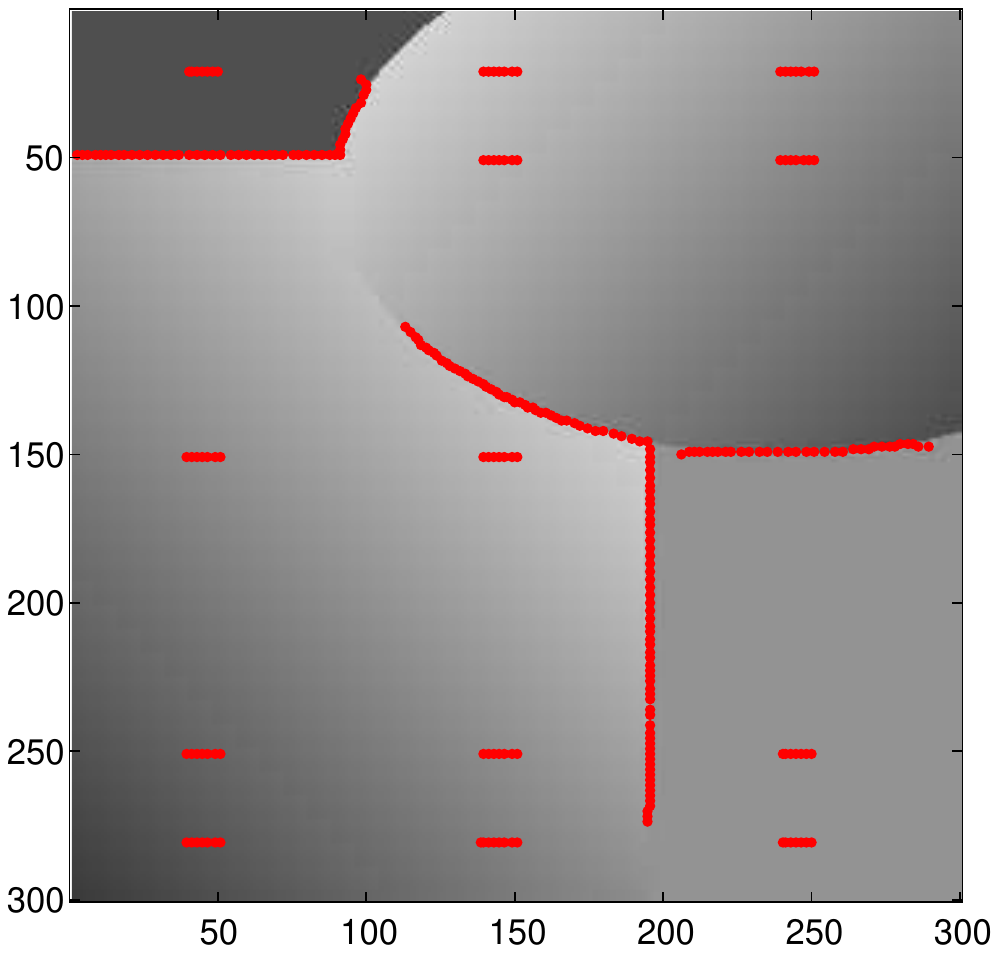}
		\includegraphics[viewport = 150 280 440 560, width=0.15\textwidth]{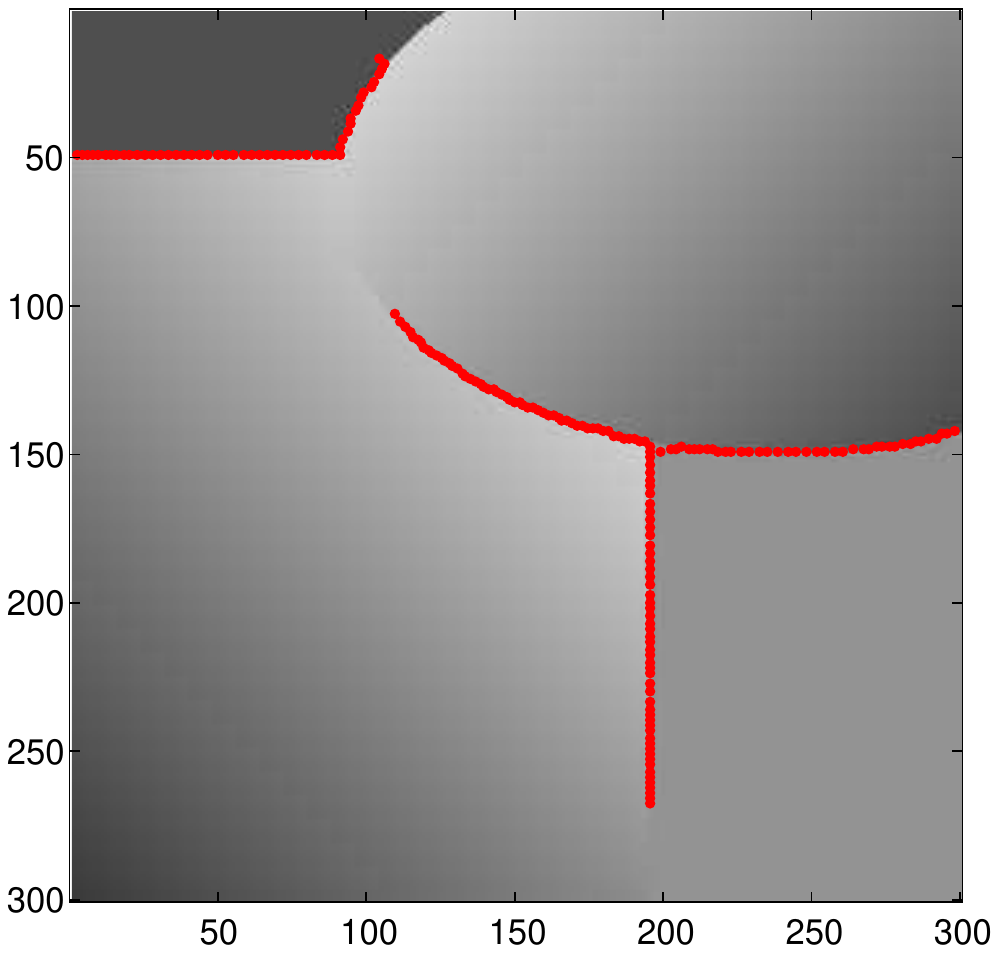}
		\includegraphics[viewport = 150 280 440 560, width=0.15\textwidth]{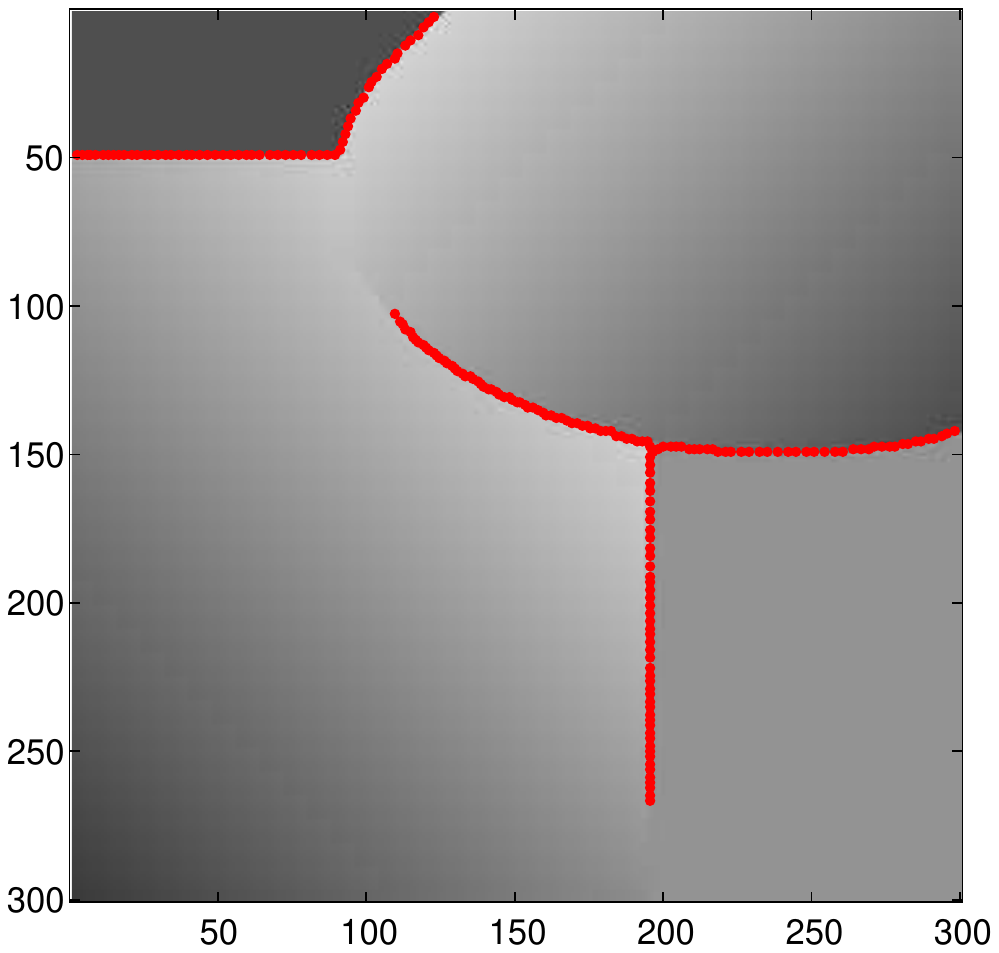}
	\caption{Image segmentation and contour detection with topology changes. The final segmentation contains two free endpoints, one triple junctions and three boundary intersection points.  Original image and contours for  $m=1, 100, 250, 400, 750, 1200$ using $\Delta t = 1$, $\sigma=1\mathrm{e}-4$,  $\lambda = 1\mathrm{e}-3$. }
	\label{fig:result_multi_line_top_change}
\end{figure}

Finally, we study an example where several topology changes occur. Figure~\ref{fig:result_multi_line_top_change} shows an example where we start with many small initial curves. A similar image is also considered in \cite{Pock2009}. Many of the small initial lines shrink and are deleted when their curve length becomes too small. Additonal topology changes occur: Near the upper left corner of the image, two curves merge at their free endpoints to one curve. Further, three free endpoints become boundary intersection points, and a triple junction emerges when a free endpoint meets another curve at an inner node. The topology changes are detected as described in Section~\ref{subsec:topology_changes} and \cite{Benninghoff2014a}.   

An advantage of the parametric method is that we can easily handle non-interface curves and complex curve networks including triple junctions. The curve evolution scheme is very similar to the scheme presented in \cite{Benninghoff2014a} for interface-curves. Instead of computing the mean value of the image function in regions, we have to solve a diffusion bulk equation. Additional to the motion of the curve in normal direction, free endpoints can move in tangential direction. 

There are alternatives to parametric methods to describe an evolving curve. The level set method \cite{OsherSethian88} is very popular for image processing applications and in particular for active contours methods, see e.g. \cite{Malladi95}, \cite{Caselles97}, \cite{Kichenassamy96}, \cite{Chan01}, \cite{Tsai01}, \cite{Sapiro06} to mention a few. In level set methods, a hypersurface is embedded as the zero level set of a function defined on the image domain $\Omega$. With level set techniques, free endpoints however cannot be handled with one single level set function: Since level set methods embed a curve as zero level set of a function $\Phi:\Omega \rightarrow \mathbb R$, the curve is an interface between two regions $\{ \Phi > 0\}$ and $\{\Phi < 0\}$. Therefore level sets are always closed or meet the image boundary at their endpoints. Non-interface curves can be handled by using two level set functions $\Phi$ and $\Psi$ and by using artificial regions, see \cite{Schaeffer13}. A curve with free endpoints can then be represented by the interface between the artificial regions $\{\Phi > 0 \} \cap \{\Psi > 0\}$ and $\{\Phi > 0 \} \cap \{\Psi < 0\}$, for example. 

Using our direct, parametric approach it is not necessary to introduce artificial regions. Further our method is very efficient, since the curve evolution is only a one-dimensional problem.

\section{Conclusion}
\label{sec:conclusion}
We proposed a new parametric approach for active contours with free endpoints. The image segmentation and denoising method presented in this article is based on a discrete version of the Mumford and Shah functional. For curves with free endpoints a flow in normal direction and a flow of the endpoints in tangential direction attracts the curves to the edges in the image. With the presented approach, we can handle both open and closed curves. The method is also suitable to be employed as postprocessing step to improve the result of a previous Chan-Vese like segmentation with interface curves and piecewise constant image approximations.

\setlength{\parskip}{0cm}

\bibliographystyle{plain}
\bibliography{literatur} 

\vfill

\end{document}